\documentclass[runningheads]{llncs}

\usepackage{eccv}

\usepackage{eccvabbrv}

\usepackage{graphicx}
\usepackage{booktabs}
\usepackage{pgfplots}
\usetikzlibrary{intersections} 
\usepgfplotslibrary{fillbetween} 
\usepackage{multirow}
\usepackage[ruled]{algorithm2e}
\usepackage{algpseudocode}

\usepackage[accsupp]{axessibility}  

\usepackage{hyperref}

\usepackage{orcidlink}

\usepackage[np, autolanguage]{numprint}  
\newcommand{\bftab}{\fontseries{b}\selectfont}
\newcommand{\abbfid}{FID$\downarrow$}
\newcommand{\abbmiou}{mIoU$\uparrow$}
\newcommand{\abbalignment}{Ali$\rightarrow$}
\newcommand{\abboverlap}{Ove$\rightarrow$}
\newcommand{\taskugen}{Un-Gen}
\newcommand{\tasktype}{Gen-Type}
\newcommand{\tasktypesize}{Gen-TypeSize}
\newcommand{\taskcompletion}{Completion}
\newcommand{\taskrefinement}{Refinement}

\begin{document}

\title{LayoutFlow: Flow Matching for \\ Layout Generation}

\author{Julian Jorge Andrade Guerreiro \inst{1}\orcidlink{0009-0002-5916-9559} \and
Naoto Inoue\inst{2}\orcidlink{0000-0002-1969-2006} \and
Kento Masui\inst{2}\orcidlink{0000-0002-4174-4378} \and \\
Mayu Otani\inst{2}\orcidlink{0000-0001-9923-2669} \and
Hideki Nakayama \inst{1}\orcidlink{0000-0001-8726-2780}}

\authorrunning{J. J. Andrade Guerreiro et al.}

\institute{The University of Tokyo, Japan\\
\email{\{guerreiro,nakayama\}@nlab.ci.i.u-tokyo.ac.jp}\\
 \and
CyberAgent, Japan\\
\email{\{inoue\_naoto,masui\_kento,otani\_mayu\}@cyberagent.co.jp}}

\maketitle

\begin{abstract}
    Finding a suitable layout represents a crucial task for diverse applications in graphic design. Motivated by simpler and smoother sampling trajectories, we explore the use of Flow Matching as an alternative to current diffusion-based layout generation models. Specifically, we propose LayoutFlow, an efficient flow-based model capable of generating high-quality layouts. Instead of progressively denoising the elements of a noisy layout, our method learns to gradually move, or flow, the elements of an initial sample until it reaches its final prediction. In addition, we employ a conditioning scheme that allows us to handle various generation tasks with varying degrees of conditioning with a single model. Empirically, LayoutFlow performs on par with state-of-the-art models while being significantly faster. 
\end{abstract}

\section{Introduction}
\label{sec:intro}
Layout design describes the process of arranging various elements, such as images, text, or other components, on a page or screen. Finding an appropriate layout forms a crucial part of creating documents, user interfaces, graphic designs, and other compositions since the arrangement of different elements can substantially impact how the intended message or purpose is communicated. Over the years, several approaches have been explored to automate layout generation in a data-driven manner using machine learning methods. Nonetheless, current layout generation models still leave room for improvement in terms of layout quality and sampling speed. 

Motivated by this observation, we propose LayoutFlow, a generative framework based on Flow Matching~\cite{lipman2023flow, liu2022flow, albergo2023building, tong2023improving} that can produce high-quality layouts while requiring less time than previous methods. Flow Matching has recently been introduced as a powerful generative framework and demonstrated strong performance on various tasks, including image generation~\cite{ma2024sit}. Intuitively, flow-based models try to learn a \textit{flow} that moves samples from a base distribution to a target distribution defined by the training data. In this work, we investigate how to apply Flow Matching for layout generation and show that flows offer a more intuitive generation process from a geometrical interpretation compared to previous models.

Most recent layout generation models\cite{chai2023layoutdm, inoue2023layoutdm, chen2024towards, zhang_layoutdiffusion, levi2023dlt, hui2023unifying, he2023diffusion-based, cheng2023play} have been based on the diffusion framework~\cite{hoDDPM} which has become the standard for various generative tasks~\cite{rombach2022high}. While diffusion models are commonly described as models that learn to gradually remove random noise that was added to the training data, the generation process can also be interpreted in analogy to flow-based models as moving samples from a Gaussian base distribution to a data distribution. In the diffusion case, however, additional randomness is introduced along the way by adding noise. Mathematically, this corresponds to a Stochastic Differential Equation (SDE), whereas a flow is defined by a simpler Ordinary Differential Equation (ODE), which we describe in more detail in \cref{sec:prelim_flow}.

\begin{figure}[t]
  \centering
  \includegraphics[width=\linewidth]{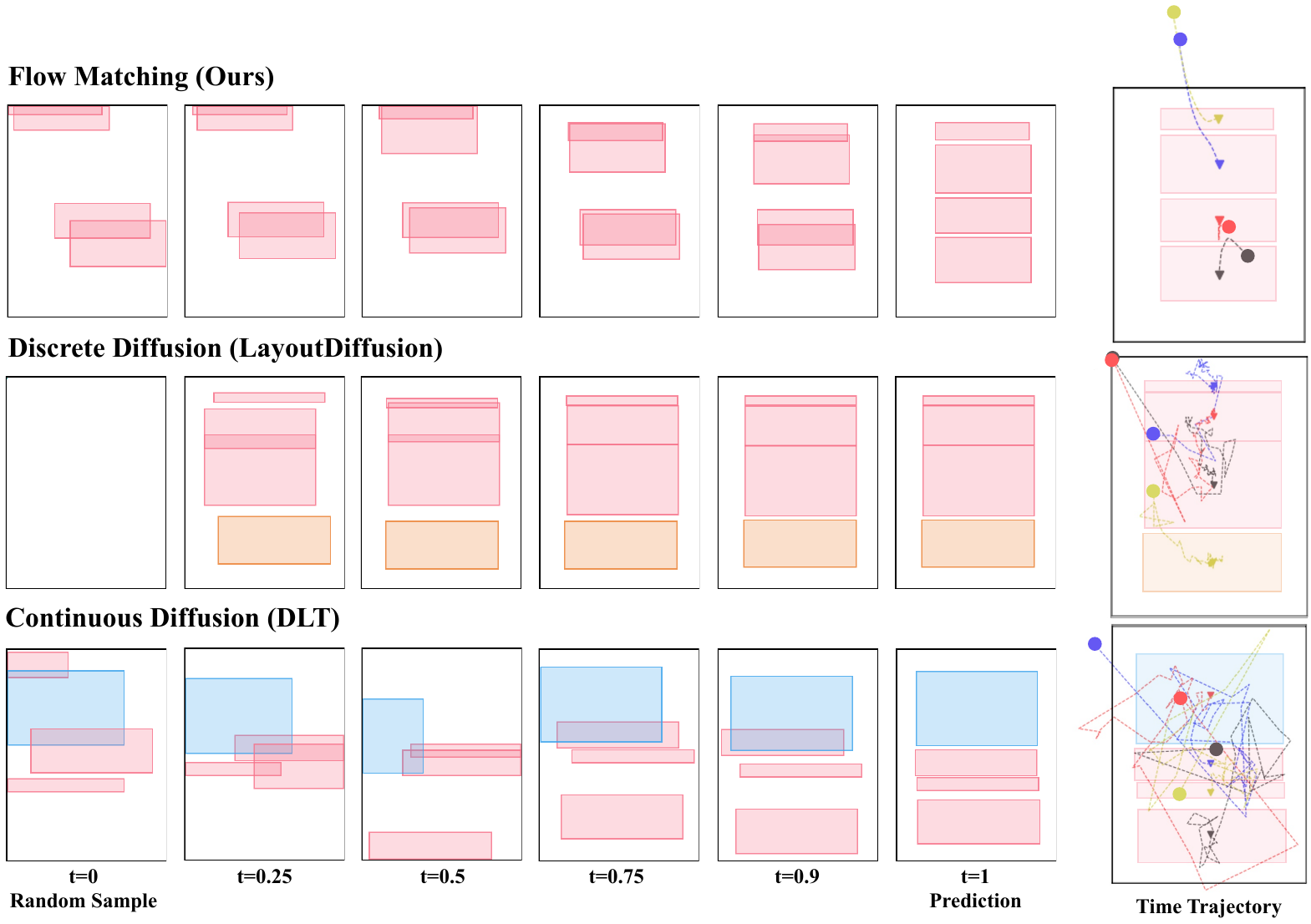}
  \caption{\textbf{Comparison of different layout generation trajectories.} Given a randomly initialized layout at time $t=0$ with fixed element sizes, we visualize different states of the generation process until the final layout at $t=1$. In addition, we overlay the trajectory, which can be interpreted as the movement of the elements over time, on top of the final layout. A circle marks the location of the initial sample and a triangle marks the final location. Flow Matching produces smooth and directed paths, whereas both diffusion models slowly converge to the final prediction under noisy trajectories with a long path length. As a result, flow-based models require fewer evaluation steps than diffusion, leading to faster sampling.     
  }
  \label{fig:comparison}
\end{figure}
In the context of layout generation, the differences between diffusion-based and flow-based models become apparent when visualizing the trajectories created by the generation process, as illustrated in \cref{fig:comparison}. The trajectories show how the sample from the base distribution is moved by the flow or backward diffusion process until it reaches the final layout prediction. While our LayoutFlow model produces smooth trajectories with short path lengths, diffusion-based models follow a noisy trajectory that constantly changes directions due to the added noise. As a result, diffusion models tend to require more sampling steps, \ie, longer inference time, than flow-based models. Altogether, we argue that learning how to move the initial sample straight toward the final prediction is much more natural than trying to do so under additional noise.   
Overall, our contributions can be summarized as follows:
\begin{enumerate}
    \item Motivated by its geometrical interpretation, we apply Flow Matching to the task of layout generation and propose a novel model called LayoutFlow, which can handle various tasks utilizing a conditioning mechanism. 
    \item On top of generating simpler trajectories, flow-based models generally also offer more choices than diffusion models, such as different prior distributions or training trajectories. We, therefore, extensively explore this additional flexibility offered by Flow Matching in the context of layout generation. 
    \item Empirically, we demonstrate that our proposed flow-based model significantly outperforms previous diffusion-based layout generation models of similar size and performs on par compared to a significantly larger model. In either case, our model greatly speeds up inference, requiring only a fraction of the time previous models need to generate samples.
\end{enumerate}

\section{Related Work}
\label{sec:relatedwork}
\subsection{Layout Generation}
Over the years, layout generation has been investigated using various methods. Early works started exploring generating layouts by minimizing an energy function based on pre-defined constraints~\cite{odonavan2014,odonovan2015}. With the advent of generative machine learning techniques, however, research shifted to a more data-centric learning approach and started to use Variational Autoencoders (VAE)~\cite{yamaguchi2021canvasvae,guo2021vinci,Arroyo_2021_VTN,lee2020n_NDN}, and Generative Adversarial Networks (GAN)~\cite{li2018layoutgan,li2020LayoutGAN,kikuchi2021_clglo}. While earlier works explored a variety of different architectures, such as Graph Convolutional Networks~\cite{lee2020n_NDN}, Long-Short Term Memories (LSTM)~\cite{guo2021vinci,Jyothi_LayoutVAE} or Transformers~\cite{yamaguchi2021canvasvae,xie2021canvasemb,gupta2021layouttransformer}, most recent research has converged to using a Transformer-based architecture due to its flexibility and strong performance.  

\textbf{Diffusion-based Models.} As diffusion models have proven to generate more diverse data and are generally easy to train, especially compared to GANs, most recent research has shifted towards exploring the diffusion process for layout generation tasks. Since layouts are represented by categorical data defining types of elements and continuous numerical values describing the element position and size, applying standard diffusion models is not straightforward. As a result, different approaches have incorporated a diffusion loss into their training process, which can be divided into continuous~\cite{hoDDPM} and discrete diffusion models~\cite{D3PM}. For the discrete case, the continuous coordinates are quantized and interpreted as different states along with the categorical element type~\cite{inoue2023layoutdm, hui2023unifying, zhang_layoutdiffusion, he2023diffusion-based}. In this discrete scenario, the forward diffusion process is modeled as a random walk between discrete states that, in addition to the quantized states, also include a mask state, which removes elements from the canvas. While LayoutDM~\cite{inoue2023layoutdm} and LDGM~\cite{hui2023unifying} rely on the mask-and-replace strategy proposed in~\cite{gu2022vqdiffusion}, LayoutDiffusion~\cite{zhang_layoutdiffusion} introduces a new mild forward process that is closer to the continuous process while still increasingly masking elements over time.

On the other hand, continuous diffusion models have been less common as they have not attained the same performance as discrete models. In particular, continuous models have struggled to produce layouts with well-aligned elements. The first continuous layout generation model based on diffusion was proposed by Chai \etal~\cite{chai2023layoutdm}, but is restricted to generating continuous element coordinates given categorical type data as a condition. Cheng \etal~\cite{cheng2023play} introduced a latent diffusion model for layout design focusing on user constraints by embedding the layout into continuous representation, on which the diffusion process is performed. The first continuous diffusion-based layout generation model acting on the layout representation and able to handle unconditional generation was proposed by Chen \etal~\cite{chen2024towards}. Their work addresses the alignment issue observed for continuous diffusion by introducing a specific regularization loss. Lastly, Levi \etal~\cite{levi2023dlt} proposed DLT, which applies a continuous diffusion process on the element coordinates and a discrete process on the element types. 

\textbf{Large Language Models.} Motivated by the strong generalization capabilities of Large Language Models (LLM), the most recent works~\cite{lin2023layoutprompter, tang2024layoutnuwa, feng2023layoutgpt} have investigated the layout generation abilities of LLMs. While LLMs provide some interesting applications, such as zero-shot synthesis, they typically consist of billions of parameters, resulting in long inference times. Moreover, current models are limited to conditional layout generation. In contrast, our proposed approach aims to offer a lightweight model for unconditional and conditional layout generation.

\subsection{Flow Matching and Diffusion Models}
Flow-based models are based on learning a mapping between samples from a simpler distribution to a data distribution. Over the years, there have been various methods, such as Normalizing Flows or Continuous Normalizing Flows, which have used neural networks to estimate such a mapping, which can subsequently be used to sample from the learned distribution~\cite{kobyzev2020normalizingflowsoverview}. Until recently, the biggest issue with flow-based models has been the need to backpropagate through an ODE during training. However, recent models have proposed training algorithms that only require explicitly solving an ODE during inference~\cite{tong2023improving, lipman2023flow, albergo2023building,liu2022flow,neklyudov2022action} demonstrating impressive results and outperforming diffusion models on the image generation task \cite{ma2024sit}. Our proposed model builds on this improvement by utilizing Flow Matching for layout generation while offering an improved geometrical interpretation.

Diffusion models, which were first proposed by Sohl-Dickenstein \etal~\cite{sohl2015deep} and later popularized by Ho \etal~\cite{hoDDPM}, can be formulated similarly to flow-based models. In fact, the term Flow Matching was inspired by the similar Score Matching loss~\cite{hyvarinen2005estimation} used by diffusion models. However, flows are characterized by an ODE, whereas the diffusion process can be formulated as an SDE with an additional stochastic component described by Brownian Motion \cite{song2020score}. As a result, diffusion models are restricted to a Gaussian as a base distribution, while flow-based models allow for more flexibility, which we also explore in our experiments. Notably, Song \etal~\cite{song2020DDIM} proposed DDIM, an alternative sampling method that makes it possible to sample using an ODE formulation on a model trained with an SDE forward process, effectively leading to smoother trajectories. In the context of layout generation, however, models have been using the SDE formulation following DDPM~\cite{hoDDPM}. Our argument for introducing Flow Matching instead of just using DDIM for smoother trajectories lies in its increased flexibility and simplicity. Moreover, if the goal is to sample using an ODE, it seems more natural to also use an ODE during training instead of the more complex SDE used in diffusion.    

\section{Preliminary: Flow Matching}
\label{sec:prelim_flow}
In this section, we first present the mathematical definition of a flow and then introduce how to train a neural network to estimate a flow via Flow Matching. The term \textit{flow} in Flow Matching refers to a mapping between samples of two distributions. In general, flow-based models aim to estimate the flow between a known source probability distribution $p_0(x)$, \eg, a Gaussian distribution, and the typically more complex data distribution $p_1(x)$. Given a data point $x \in \mathbb{R}^d$, the flow $\phi: [0, 1] \times \mathbb{R}^d \rightarrow \mathbb{R}^d$ can be defined by the following Ordinary Differential Equation (ODE)
\begin{align}    
    \frac{d}{dt} \phi_t (x) = v_t (\phi_t(x)), \quad \phi_0(x)=x_0,
    \label{eq:flowode}
\end{align}
where the time-dependent vector field $v_t: [0, 1] \times \mathbb{R}^d \rightarrow \mathbb{R}^d$ is said to construct the flow~\cite{lipman2023flow} with the boundary condition $\phi_0(x)=x_0$. In other words, the flow $\phi_t(x)$ describes how an initial sample $x_0$ is transported over time and is given as the solution to the ODE in  \cref{eq:flowode}. As the flow transports samples with respect to time, the associated probability distribution is transformed according to the change of variables formula and creates a probability density path 
\begin{align}    
    p_t = [\phi_t]_{*} p_{0} = p_0 ( \phi_t^{-1}(x))\det \left[  \frac{d\phi_t^{-1}}{dt} (x) \right]
    \label{eq:pushforward}
\end{align}
with $\phi_t^{-1}$ denoting the inverse of the flow and $*$ representing the push-forward operator. If the flow constructed by the vector field $v_t$ satisfies \cref{eq:pushforward}, it is said that $v_t$ generates the probability path $p_t$. This is the case if $v_t$ follows the continuity equation, which holds for our considerations in this paper.

Instead of estimating the flow directly, Flow Matching trains a neural network $u_{\theta}(t, x)$ with weights $\theta$ to match the vector field $v_t$ and then solves the ODE for a given boundary condition $x_0$ to obtain the flow. The Flow Matching loss function is defined as the expectation over uniformly sampled timesteps and samples along the probability density path as follows
\begin{align}
    \mathcal{L}_{FM}(\theta) = \mathbb{E}_{t \sim 
    \mathcal{U}(0,1),x \sim p_t({x})} \| u_{\theta}(t,x) - v_t (x) \|^2.
    \label{eq:FMloss}
\end{align}
However, this loss alone does not allow us to train the model as we usually do not have access to $p_t$ nor $v_t$. As an extension, Lipman \etal~\cite{lipman2023flow} have proposed a Conditional Flow Matching objective and is given as  
\begin{align}
    \mathcal{L}_{CFM}(\theta) = \mathbb{E}_{t \sim \mathcal{U} (0,1), z \sim q(z), x \sim p_t({x|z})} \| u_{\theta}(t,x) - v_t (x|z) \|^2,
    \label{eq:CFMloss}
\end{align}
with $v_t(x|z)$ denoting a conditional vector field generating the conditional probability path $p_t(x|z)$ and $q(z)$ being a distribution over a latent conditioning variable $z$. By introducing the conditional formulation of the Flow Matching objective, training a neural network becomes possible by choosing and training with a conditional vector field and its associated probability path. For an overview of the various choices and more details on Conditional Flow Matching, we recommend~\cite{lipman2023flow, tong2023improving}.  

\section{Conditional Flow Matching for Layout Generation}
\subsection{Data Representation}
A layout typically consists of various elements, such as text, tables, or images placed on a canvas. Mathematically, we can describe a layout $\mathcal{A}$ as a set of $N$ elements $\mathbf{e}^k$ with $k \in [ 1, N]$ in the following way
\begin{align}
\mathcal{A} = \{ \mathbf{e}^k = (\mathbf{g}^k, \mathbf{a}^k) | k \in [1, N] \},
\end{align}
where we split each element into its geometrical information $\mathbf{g}^k = (c_x^k, c_y^k, w^k, h^k)$ described by the bounding box center $(c^k_x, c^k_y)$, width $w^k$ and height $h^k$, and its category information $\mathbf{a}^k$ describing the element type, such as text or image.  In our considerations, the elements on the canvas do not possess an inherent order, which is why the canvas is considered a set of elements. 
Layout generation can be divided into various tasks depending on the conditioning assumptions. For example, in the unconditional generation setting, it is assumed that no prior knowledge is available, while for the refinement task, a complete but coarse layout is given. 
Since Flow-Matching requires the input data to be continuous, we convert the categorical information $\mathbf{a}^k$ into a continuous embedding, denoted as $\Tilde{\mathbf{a}}^k$, using $B$ Analog Bits \cite{chen2023analog}. To learn the flow, we define a data sample $\mathbf{x} \in \mathbb{R}^{(4+B) \cdot  N_{max}}$ as the concatenation of all the elements $\mathbf{g}^k$ and $\Tilde{\mathbf{a}}^k$ in a single layout. If the number of elements is smaller than $N_{max}$, the maximum number of elements contained in a single layout across the entire dataset, we pad the remaining dimensions.     

\subsection{Training}

\begin{algorithm}[t]
    \caption{Training}\label{alg:training}
    \DontPrintSemicolon
        $\mathbf{x}_0 \sim p_0(\mathbf{x})$, $\mathbf{x}_1 \sim p_{data}(\mathbf{x})$, $t \sim \mathcal{U}(0,1)$ \tcp*{Sampling from distributions}
        
        $\mathbf{x}_t \gets (1-t)\mathbf{x}_0 + t \mathbf{x}_1$ \tcp*{Linear interpolation}
        
        $\mathbf{v}_t \gets \mathbf{x}_1 - \mathbf{x}_0 $ \tcp*{Calculate the constant vector field}
        
        $\mathbf{u}_t \gets f_{\theta}(\mathbf{x}_t, t) $ \tcp*{Get vector field prediction from network}
        
        $ \mathcal{L}(\theta) = \| \mathbf{u}_t - \mathbf{v}_t \|^2 + \lambda \| \mathbf{u}^{\mathbf{g}}_\theta (t, \mathbf{x_t}) - (\mathbf{g_1} - \mathbf{g_0}) \|_1 $ \tcp*{Compute loss} 
\end{algorithm}

\begin{figure}[t]
  \centering
  \includegraphics[width=\linewidth]{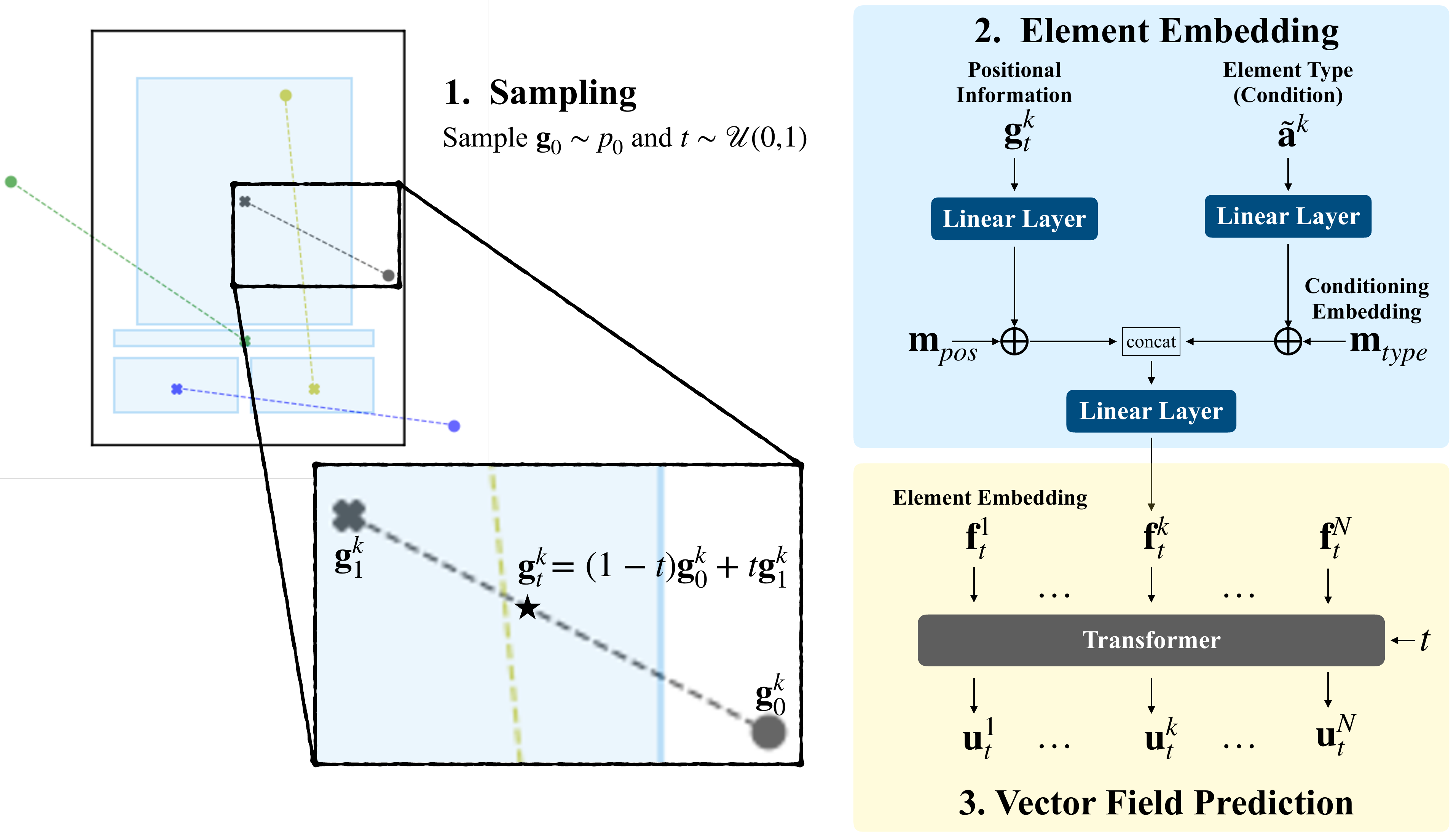}
  \caption{\textbf{Overview of the training procedure of LayoutFlow for the type-conditioned scenario.} First, we sample an initial layout from a base distribution and a time $t$. Then, an intermediate sample $\mathbf{g}_t$ is calculated by linearly interpolating between the initial sample and the ground truth layout. Each intermediate element is embedded jointly with the given element condition $\Tilde{\mathbf{a}}^k$. Lastly, the Transformer architecture takes all the element embeddings to predict a vector field.     
  }
  \label{fig:training}
\end{figure}

To generate layouts, we want to find a way to sample from a data distribution $p_{\text{data}}(\mathbf{x})$, whereas we only have access to a limited amount of samples $\mathbf{x}$ through our training data. More specifically, our goal is to train a neural network that allows us to sample from the underlying data distribution and guide the sampling based on certain conditional constraints. To that end, we propose LayoutFlow, a layout generation model based on Flow Matching.  

An overview of the training procedure for the type-conditioned scenario is illustrated in \cref{fig:training} and summarized in \cref{alg:training} for the unconditional case. In the first step, we randomly sample a layout $\mathbf{x}_0$ from a prior distribution $p_0$, \eg, a Gaussian distribution. This sample will typically not be well-aligned, and some elements might not be located on the canvas. In addition, we also randomly select a time $t$, which determines the intermediate sample $\mathbf{x}_t$ on the flow trajectory used during training.  
For LayoutFlow, we train on a simple linear trajectory following~\cite{tong2023improving,lipman2023flow}. Therefore, the intermediate sample is simply obtained by linearly interpolating between the ground truth layout $\mathbf{x}_1$ and the initial sample $\mathbf{x}_0$. The same trajectory is not used again during training since $\mathbf{x}_0$ and $t$ change with every iteration.

The network $f_{\theta}$ is trained to output the conditional vector field, corresponding to the derivative of the trajectory as defined in \cref{eq:flowode}. This derivative is simply a constant for the linear trajectory case, specifically the difference between $\mathbf{x}_0$ and $\mathbf{x}_1$. Intuitively, the network learns to output a direction pointing toward a data sample, which can be used during sampling to move the initial data samples toward a better prediction. While we train our network using linear trajectories, this does not mean that the model will also produce straight trajectories during sampling. As mentioned in \cref{eq:CFMloss}, the linear trajectory, which is derived by the conditional vector field in \cref{eq:CFMloss}, acts as a proxy to learn the actual vector field in \cref{eq:FMloss} defined by Flow Matching.     

In our experiments, we found that using only the Mean Squared Error (MSE) as a loss function tends to lead to poorly aligned layouts, which results in suboptimal perceptual quality. This observation is similar to the issue described in~\cite{chen2024towards}. We hypothesize that this can be attributed to the fact that the MSE penalizes all mistakes within the same distance equally, regardless of direction, whereas the alignment between elements is usually only measured along the horizontal and vertical direction. As a result, the minimization objective does not fully align with the perceptual quality. To tackle this issue, we add an $L_1$ regularization on the geometrical part of the output $u^{\mathbf{g}}_\theta$ as the $L_1$ loss encourages sparsity of the error vector, which can also be interpreted as minimizing the error along the axis dimensions. Therefore, our final training loss can be written as
\begin{align}
    \mathcal{L}(\theta) = \mathcal{L}_{\mathrm{CFM}} (\theta) + \lambda \mathcal{L}_{1} (\theta) = \mathcal{L}_{\mathrm{CFM}} (\theta) + \lambda \| u^{\mathbf{g}}_\theta (t, \mathbf{x_t}) - (\mathbf{g_1} - \mathbf{g_0})\|_1,
    \label{eq:loss}
\end{align}
where $\lambda$ is a hyperparameter.

\begin{figure}[t]
    \centering
    \begin{minipage}[b]{0.55\textwidth}
        \centering
        \includegraphics[width=\linewidth]{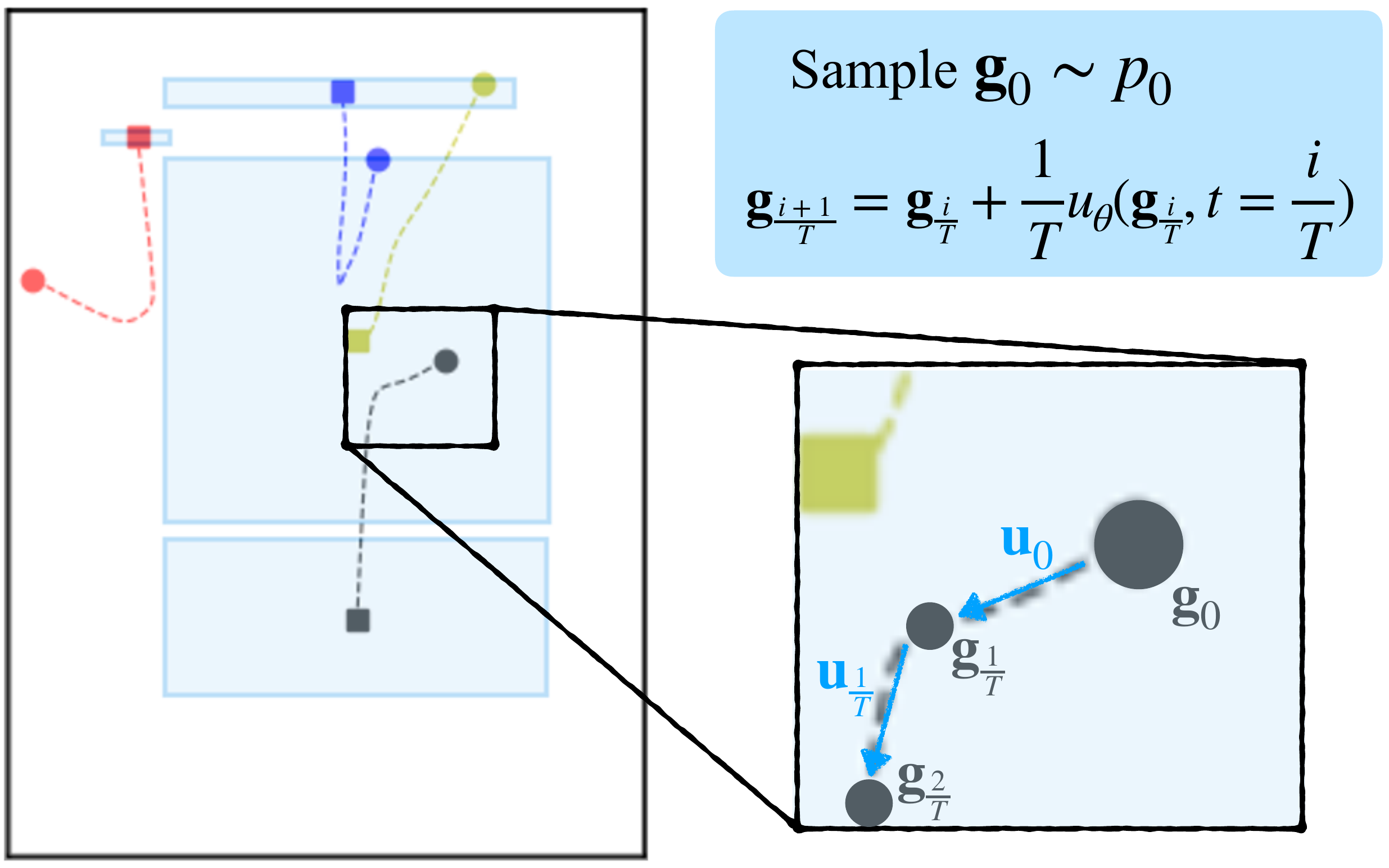}
        \caption{\textbf{Overview of the type-conditioned inference process.} The initial sample is autoregressively moved in the predicted direction.}
        \label{fig:sampling}
    \end{minipage}
    \hspace{0.25cm}
    \begin{minipage}[b]{0.35\textwidth}
        \centering
        \includegraphics[width=0.9\linewidth]{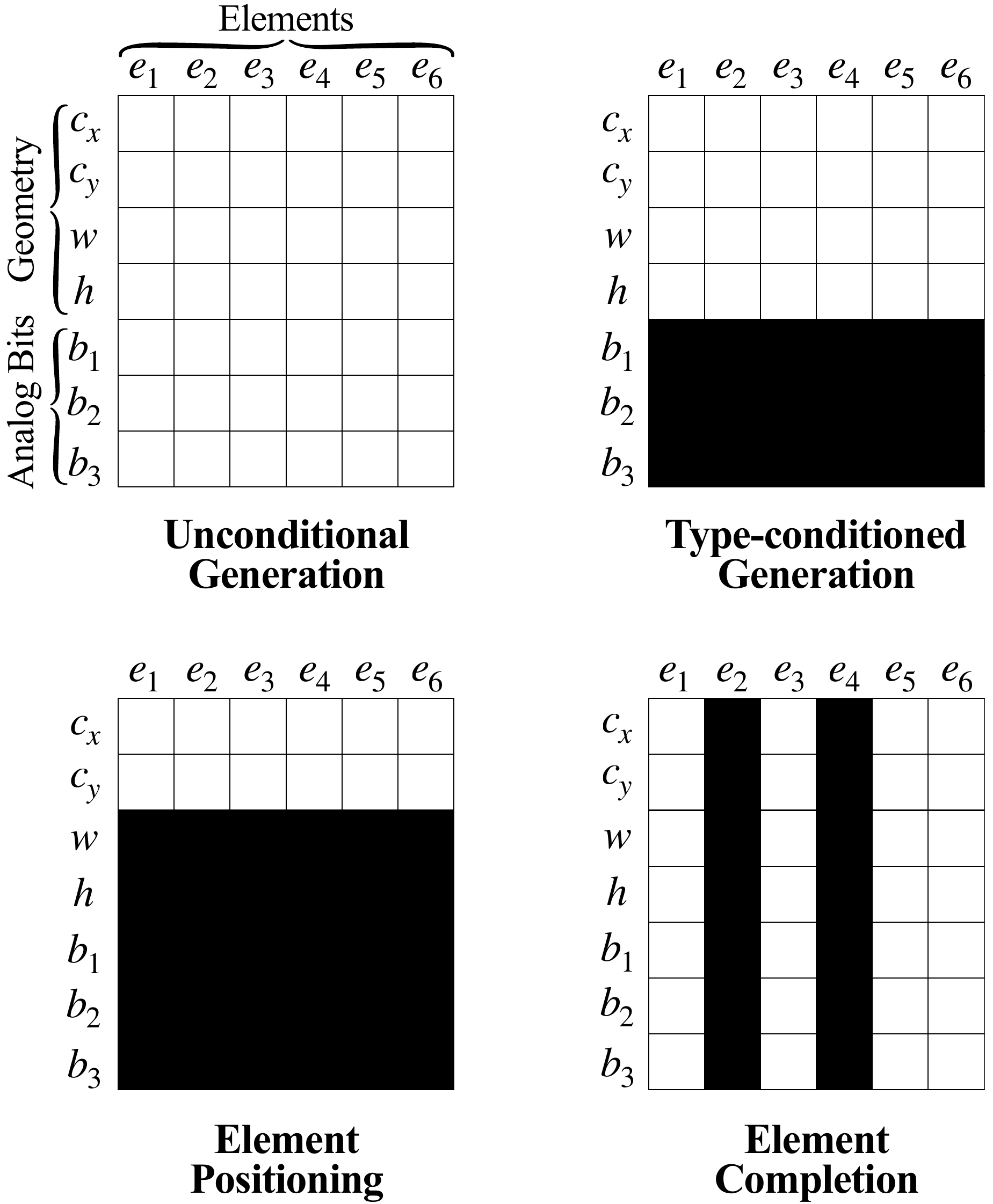}
        \caption{\textbf{Condition masks.} Black indicates parts given as conditions.}
        \label{fig:masking}
    \end{minipage}
\end{figure}

\subsection{Sampling}
After training our neural network to predict a vector field, we can generate new layouts by sampling from our initial distribution and solving the ODE describing the flow as defined in \cref{eq:flowode}. In practice, this can be done using any numerical ODE solver. \Cref{fig:sampling} illustrates the sampling process using the Euler method~\cite{euler1794institutiones} to solve the ODE. For a fixed number of evaluation steps $T$, the initial sample $x_0$ is moved along the direction predicted by the network in an autoregressive manner. Therefore, the backward process can be denoted as   
\begin{align}
    \mathbf{x}_{\frac{i+1}{T}} = \mathbf{x}_{\frac{i}{T}} + \frac{1}{T} u_{\theta} \left( \frac{i}{T}, \mathbf{x}_{\frac{i}{T}} \right),
\end{align}
with $i \in \mathbb{N}$  and $i \in [0, T-1]$ indicating how far along the sample has been moved. 

\subsection{Network Architecture}
Here, we describe the actual implementation of the network $f_{\theta}$ to efficiently handle the layout.
We first embed the elements of the sampled intermediate layout $\mathbf{x}_t$ and then use a Transformer to predict the conditional vector field as depicted in \cref{fig:training}. First, we separately embed the geometrical information $\mathbf{g}^k_t$ and the element type $\Tilde{\mathbf{a}}^k_t$ using a linear layer. A conditioning mask is embedded and added to the geometrical and type embedding to inform the network about which parts of the input act as a condition. The example in \cref{fig:training} illustrates this for the type-conditioned case, where the network is informed through the masks $\mathbf{m}_{type}$ and $\mathbf{m}_{pos}$ that $\Tilde{\mathbf{a}}^k$ is taken from $\mathbf{x}_1$ while only the geometrical information $\mathbf{g}^k_t$ was taken from the training trajectory. In a subsequent step, we concatenate both embeddings and pass them through another linear layer to obtain a high-dimensional embedding $\mathbf{f}_t^k$ representing an element of the intermediate layout $\mathbf{x}_t$. Next, all element embeddings $\mathbf{f}_t^k$ are processed by a transformer encoder network together with the temporal information $t$ incorporated in each layer. Finally, the output of the Transformer is projected back to the same dimension as the input layout using a linear layer representing the vector field.  

\subsection{Conditioning Mechanism}
As layout generation involves various tasks with different assumptions about available knowledge, we employ a conditioning mechanism to perform these tasks with just a single model. Similarly to~\cite{chen2024towards, hui2023unifying, levi2023dlt}, we mark the conditioning input by including the information in the element embedding while distinguishing between the four different scenarios depicted in \cref{fig:masking}. During training, we simply randomly sample each mask with the same probability.  Alternatively, it is also possible to only train an unconditional model and impose the conditions during sampling through a guidance strategy, similar to~\cite{inoue2023layoutdm, zhang_layoutdiffusion}. However,  we empirically found that including conditioning during training works better. For a more detailed analysis, we refer to the supplementary material.         

\section{Experiments}
To compare LayoutFlow with other methods, we first perform a series of experiments on various layout generation tasks. In addition, we conduct an ablation study on our proposed method. 

\subsection{Experimental Settings}

\textbf{Datasets.}  We evaluate the performance of our network on the RICO~\cite{deka2017rico} and PubLayNet~\cite{zhong2019publaynet} datasets, following previous methods. RICO consists of over 66k User Interface (UI) layouts with 25 element categories, while PubLayNet comprises over 360k document layouts annotated with 5 different element types. We train our network using the dataset split described in~\cite{zhang_layoutdiffusion, jiang2023layoutformer++}, which discards layouts containing more than 20 elements.

\textbf{Implementation Details.} Similar to LayoutDM and DLT, we employ a 4-layer Transformer network with 8 attention heads and a model dimension of 512 for a fair comparison. The model is trained using the AdamW optimizer with a learning rate of 0.0005. In addition to the MSE Flow Matching loss, we regularize the training by calculating the $L_1$ loss between the geometrical information as described in \cref{eq:loss} with $\lambda=0.2$. For inference, we obtain the flow using a NeuralODE solver~\cite{chen2018neuralode} for 100 steps with the Euler method.

\textbf{Evaluation Metrics.} Following previous work~\cite{zhang_layoutdiffusion, chai2023layoutdm, levi2023dlt, chen2024towards}, we evaluate our method as well as comparable approaches using \textit{Frechet Inception Distance (FID)}~\cite{heusel2017fid}, \textit{Maximum Intersection over Union (mIoU)}~\cite{kikuchi2021_clglo}, \textit{Alignment}~\cite{lee2020n_NDN}, and \textit{Overlap}~\cite{li2018layoutgan}. \textit{FID} and \textit{mIoU} inherently evaluate both fidelity and diversity~\cite{heusel2017gans} of the generated results. For the \textit{FID} calculation, we use the same network with identical weights as LayoutDiffusion~\cite{zhang_layoutdiffusion} to measure the similarity between the generated layouts and the original dataset based on the feature space. The \textit{mIoU} metric compares sets of layouts and measures their similarity by optimally matching generated and real layouts that maximize the average \textit{IoU}. On the other hand, \textit{Alignment} and \textit{Overlap} capture the fidelity property. Nonetheless, both metrics should be judged with respect to a reference dataset, where the best outputs lead to similar values. 

\textbf{Comparison with Existing Approaches.} To validate the performance of our proposed method, we compare it to state-of-the-art techniques. Since there is no standard way of splitting the datasets, we need to retrain the models that do not follow the dataset split used in LayoutDiffusion~\cite{zhang_layoutdiffusion} for a fair comparison. This fact limits us to only compare to methods that have made their code publicly available. Therefore, we compare our method against the two discrete diffusion models LayoutDM~\cite{chai2023layoutdm} and LayoutDiffusion~\cite{zhang_layoutdiffusion} as well as the discrete-continuous diffusion model DLT~\cite{levi2023dlt}. Where applicable, we additionally compare LayoutFlow to the non-diffusion-based models LayoutTransformer~\cite{gupta2021layouttransformer}, LayoutFormer\texttt{\char`+\char`+}~\cite{jiang2023layoutformer++}, NDN-none~\cite{lee2020n_NDN} and RUITE~\cite{rahman2021ruite}.

\textbf{Tasks.} We test LayoutFlow and existing approaches on various common unconditional and conditional layout generation tasks. \noindent\emph{\taskugen} describes the layout generation task without any constraints. For the conditional scenario, we consider \noindent\emph{\tasktype} as assuming the element types as given, while \noindent\emph{\tasktypesize} assumes element types and sizes are known. Lastly, we consider the \noindent\emph{\taskcompletion} task, which takes in a partial layout that is missing up to 20\% of its elements, and the \noindent\emph{\taskrefinement} task following \cite{zhang_layoutdiffusion} with a standard deviation of 0.01. 

\begin{table}[t]
    \centering
    \caption{\textbf{Quantitative results for various layout generation tasks on the RICO and PubLayNet datasets.} 
    The two best results are highlighted in \textbf{bold} and \underline{underlined}. The $\rightarrow$ symbol indicates best results are the ones closest to the validation data. Models marked with * have been retrained.}
    \resizebox{\columnwidth}{!}{
    \begin{tabular}{llrrrrrrrrr}
        \toprule
        & &  \multicolumn{4}{c}{RICO} &&  \multicolumn{4}{c}{PubLayNet}\\
        Task & Model                                    &  \abbfid          & \abbalignment     & \abboverlap       & \abbmiou          & $\quad$ & \abbfid & \abbalignment     & \abboverlap       & \abbmiou \\
        \midrule
        \multirow{6}{*}{\taskugen}& LayoutTransformer   & 24.32             & 0.037             & 0.542             & 0.587             && 30.05            & 0.067             & 0.005             & 0.359\\
        & LayoutFormer\texttt{\char`+\char`+}           & 20.20             & \underline{0.051} & 0.546             & \bftab{0.634}     && 47.08            & 0.228             & \underline{0.001} & 0.401\\
        \cmidrule{2-11} 
        & LayoutDM*                                     & 4.43              & 0.143             & 0.584             & 0.582             && 36.85            & 0.180             & 0.132             & 0.382\\
        & DLT*                                          & 13.02             & 0.271             & 0.571             & 0.566             && 12.70            & 0.117             & 0.036             & \bftab{0.431}\\
        & LayoutDiffusion                               & \underline{2.49}  & \bftab{0.069}     & \underline{0.502} & \underline{0.620} && \bftab{8.63}     & \underline{0.065} & \bftab{0.003}     & 0.417\\
        & LayoutFlow (ours)                             & \bftab{2.37}      & 0.150             & \bftab{0.498}     & 0.570             && \underline{8.87} & \bftab{0.057}     & 0.009             & \underline{0.424}\\ 
        \midrule
        \multirow{6}{*}{\tasktype} & NDN-none           &  13.76            & 0.560             &  0.550            & \underline{0.350} &&  35.67           &  0.350            &  0.170            & 0.310\\
        & LayoutFormer\texttt{\char`+\char`+}           &  2.48             & \underline{0.124} &  0.537            & \bftab{0.377}     &&  10.15           &  \bftab{0.025}    &  \underline{0.009}& 0.333\\
        \cmidrule{2-11} 
        & LayoutDM*                                     &  2.39             & 0.222             &  0.598            & 0.341             &&  39.12           &  0.267            &  0.139            &  0.348\\
        & DLT*                                          &  6.64             & 0.303             &  0.616            &  0.326            &&  7.09            &  0.097            &  0.040            & \underline{0.349}\\
        & LayoutDiffusion                               &  \underline{1.56} & \bftab{0.124}     &  \bftab{0.491}    &  0.345            &&  \underline{3.73}&  \underline{0.029}&  \bftab{0.005}    & 0.343\\
        & LayoutFlow (ours)                             & \bftab{1.48}      & 0.176             & \underline{0.517} &  0.322            && \bftab{3.66}     & 0.037             & 0.011             &\bftab{0.350}\\ 
        \midrule
        \multirow{3}{*}{\tasktypesize} & LayoutDM*      &  \underline{1.76} &  \bftab{0.175}    &  \underline{0.606}&  0.424            &&  29.91           &  0.246            &  0.160            &\underline{0.436}\\
        & DLT*                                          &  6.27             &  0.332            &  0.609            &  0.424            && \underline{5.35} & \underline{0.130} & \underline{0.053} & 0.426\\
        & LayoutFlow (ours)                             &  \bftab{1.03}     & \underline{0.283} &  \bftab{0.523}    & \bftab{0.470}     &&  \bftab{1.26}    &  \bftab{0.041}    &  \bftab{0.031}    & \bftab{0.454}\\ 
        \midrule
        & \textbf{Validation Data}                      & 2.10              & 0.093             & 0.466             &  0.658            && 8.10             & 0.022             & 0.003             & 0.434\\
        \bottomrule
    \end{tabular}
    }
    \label{tab:quant_results}
\end{table}

\subsection{Evaluation}
\textbf{Quantitative Evaluation.} In \cref{tab:quant_results}, we report the results of our experiments comparing LayoutFlow with existing approaches for different layout generation tasks. In terms of \textit{FID}, our model outperforms state-of-the-art methods across all three tasks, except for a close second place on the PubLayNet \textit{\taskugen}~task, proving its strong capabilities to model the underlying sample distribution. In particular, the \textit{FID} score of 2.37 produced by LayoutFlow on the unconditional generation for RICO almost matches the score obtained when comparing the validation with the test dataset. Note that for the conditional tasks, the \textit{FID} score can become even lower than the validation dataset since the generated images share more similarities through conditioning. Regarding the geometrical metrics \textit{Alignment}, \textit{Overlap}, and \textit{mIoU}, our method consistently outperforms LayoutDM and DLT while usually producing values close to LayoutDiffusion, a significantly larger model with 85M parameters compared to around 15M used by LayoutFlow. A comparison between the quality and the sampling speed is illustrated in \cref{fig:performance_time}, clearly showing that our method closes the gap between inference speed and performance. While non-diffusion models outperform diffusion models on some geometrical metrics, the drastically higher \textit{FID} score implies that the generated results lack diversity.

On the completion task, shown in \cref{tab:compl_refinement}, LayoutFlow is able to clearly outperform LayoutDM, the only diffusion model that can handle that task, except for falling behind on \textit{Alignment} for RICO. On the refinement task, LayoutFlow significantly improves the noisy layouts on PubLayNet, while being the second-best model for RICO, underlining its broad capabilities.  

\begin{table}[t]
    \centering
    \caption{\textbf{Quantitative results for the completion and refinement tasks on the RICO and PubLayNet datasets.} 
    The two best results are highlighted in \textbf{bold} and \underline{underlined}. Models marked with * have been retrained.}
    \resizebox{\columnwidth}{!}{
    \begin{tabular}{llrrrrrrrrr}
        \toprule
        & &  \multicolumn{4}{c}{RICO} &&  \multicolumn{4}{c}{PubLayNet}\\
        Task & Model &  \abbfid & \abbalignment & \abboverlap & \abbmiou & $\quad$ & \abbfid & \abbalignment & \abboverlap & \abbmiou \\
        \midrule
        \multirow{2}{*}{\taskcompletion}        
        & LayoutDM*                             & 6.80              &  \bftab{0.054}    &  0.630            & 0.678             && 25.02            & 0.169             & 0.107             & 0.678\\
        & LayoutFlow (ours)                     & \bftab{1.51}      &  0.150            &  \bftab{0.474}    & \bftab{0.741}     && \bftab{1.10}     & \bftab{0.054}     & \bftab{0.127}     & \bftab{0.746}\\ 
        \midrule
        \multirow{5}{*}{\taskrefinement}        
        & RUITE                                 & 7.93              & 0.177             & 0.492             & 0.658             && 7.89             & 0.073             & 0.038             & 0.637\\
        & LayoutFormer\texttt{\char`+\char`+}   & 3.67              & \underline{0.141} & 0.503             & 0.656             && 2.94             & 0.042             & 0.013             & 0.642\\
        \cmidrule{2-11} 
        & LayoutDM*                             & 2.91              & 0.143             & 0.575             & 0.437             && 48.61            & 0.286             &  0.173            & 0.372\\
        & LayoutDiffusion                       & \bftab{0.55}      &  \bftab{0.102}    & \bftab{0.469}     & \bftab{0.719}     && \underline{2.05} & \underline{0.035} & \underline{0.008} & \underline{0.660}\\
        & LayoutFlow (ours)                     & \underline{0.77}  &  0.152            & \underline{0.455} & \underline{0.700} && \bftab{0.18}     & \bftab{0.020}     & \bftab{0.005}     & \bftab{0.723}\\ 
        \midrule
        & \textbf{Validation Data}              & 2.10              & 0.093             & 0.466             &  0.658            && 8.10             & 0.022             & 0.003             &0.434\\
        \bottomrule
    \end{tabular}
    }
    \label{tab:compl_refinement}
\end{table}

\begin{figure}[t]
    \centering
    \begin{minipage}[b]{0.48\textwidth}
        \centering
        \resizebox{\textwidth}{!}{
        \begin{tikzpicture}
    \begin{axis}[
        xlabel={Inference time per sample (in seconds)},
        ylabel={FID score},
        y label style={yshift=-14pt},
        xmode=log,
        xmin=0.0001, xmax=10,
        ymin=0, ymax=15,
        xtick={0.0001, 0.001, 0.01, 0.1, 1, 10},
        ytick={0, 2, 4, 6, 8, 10, 12, 14},
        legend pos=north east,
        legend style={cells={anchor=west}}, 
        xmajorgrids=true,
        xminorgrids=true,
        ymajorgrids=true,
        yminorgrids=true,
        grid style=dashed,
        ]
    \addplot[
        only marks,
        color=red,
        mark=pentagon*,
        ]
        coordinates {
        (1.61, 2.45)
        };
        \addlegendentry{LayoutDiffusion}
    \addplot[
        only marks,
        color=green,
        mark=triangle*,
        ]
        coordinates {
        (0.00035, 13.02)
        };
        \addlegendentry{DLT}
    \addplot[
        only marks,
        color=magenta,
        mark=diamond*,
        ]
        coordinates {
        (0.00166, 4.43)
        };
        \addlegendentry{LayoutDM}
    \addplot[
        only marks,
        color=blue,
        mark=square*,
        ]
        coordinates {
        (0.000175, 2.37)
        };
        \addlegendentry{LayoutFlow(ours)}

    \end{axis}
\end{tikzpicture}
        }
        \caption{\textbf{Quality-Speed Comparison} }
        \label{fig:performance_time}
    \end{minipage}
    \begin{minipage}[b]{0.48\textwidth}
        \centering
        \includegraphics[width=\textwidth]{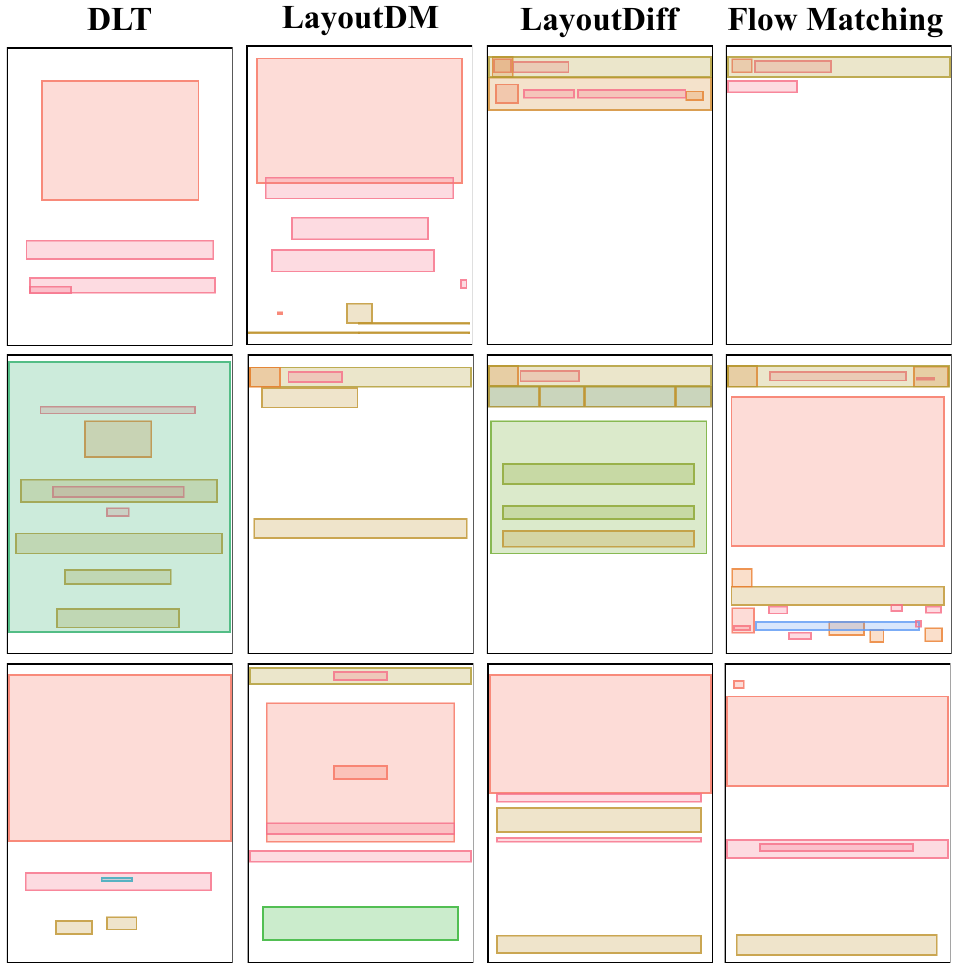}
        \caption{\textbf{Examples for \taskugen}}
        \label{fig:qual_uncond}
    \end{minipage}
\end{figure}

\begin{figure}[t]
    \centering
    \begin{subfigure}[b]{0.57\textwidth}
        \centering
        \includegraphics[height=4.5cm]{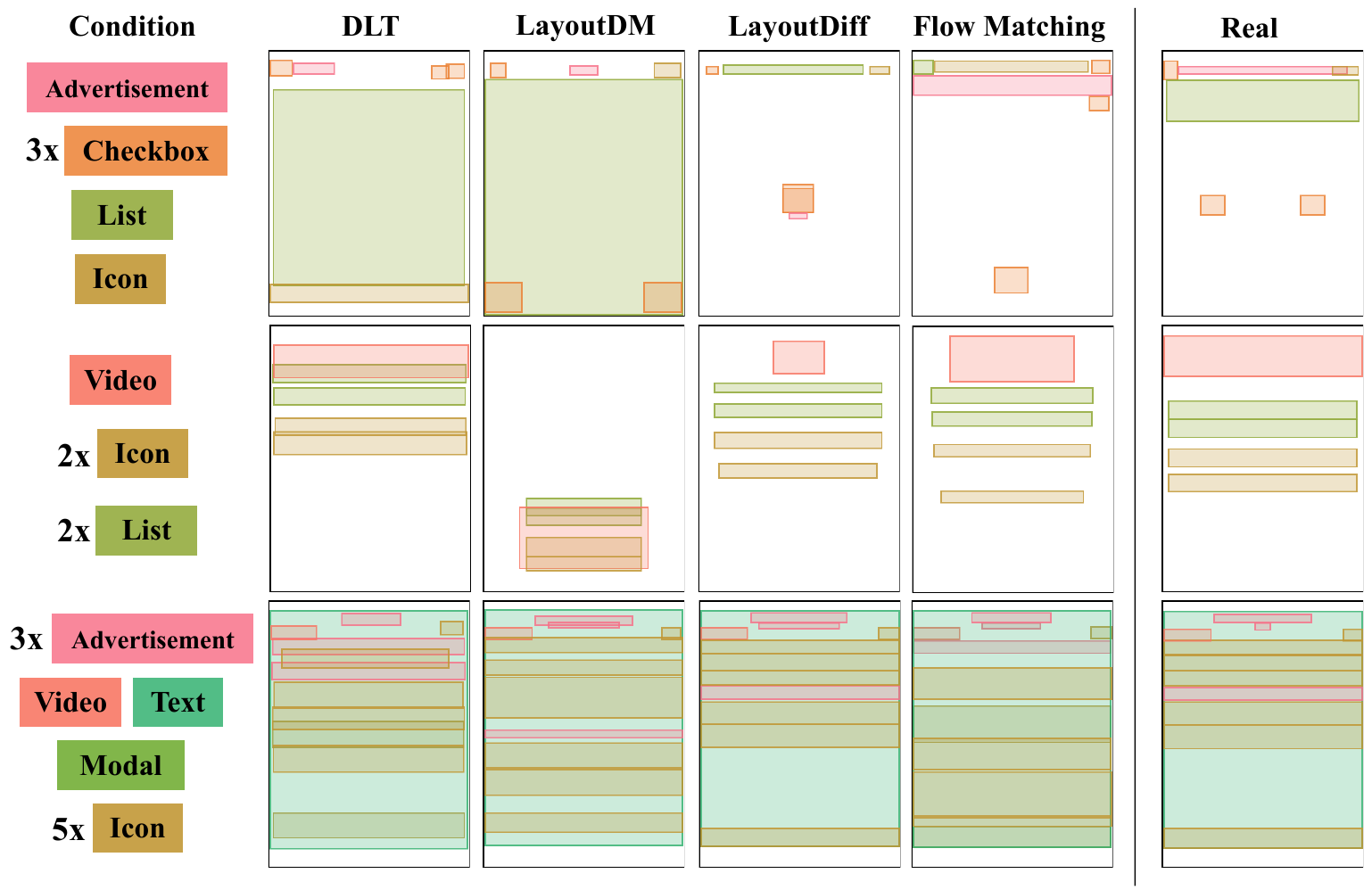}
        \caption{\textbf{Examples for \tasktype}}
    \end{subfigure}
    \begin{subfigure}[b]{0.4\textwidth}
        \centering
        \includegraphics[height=4.5cm]{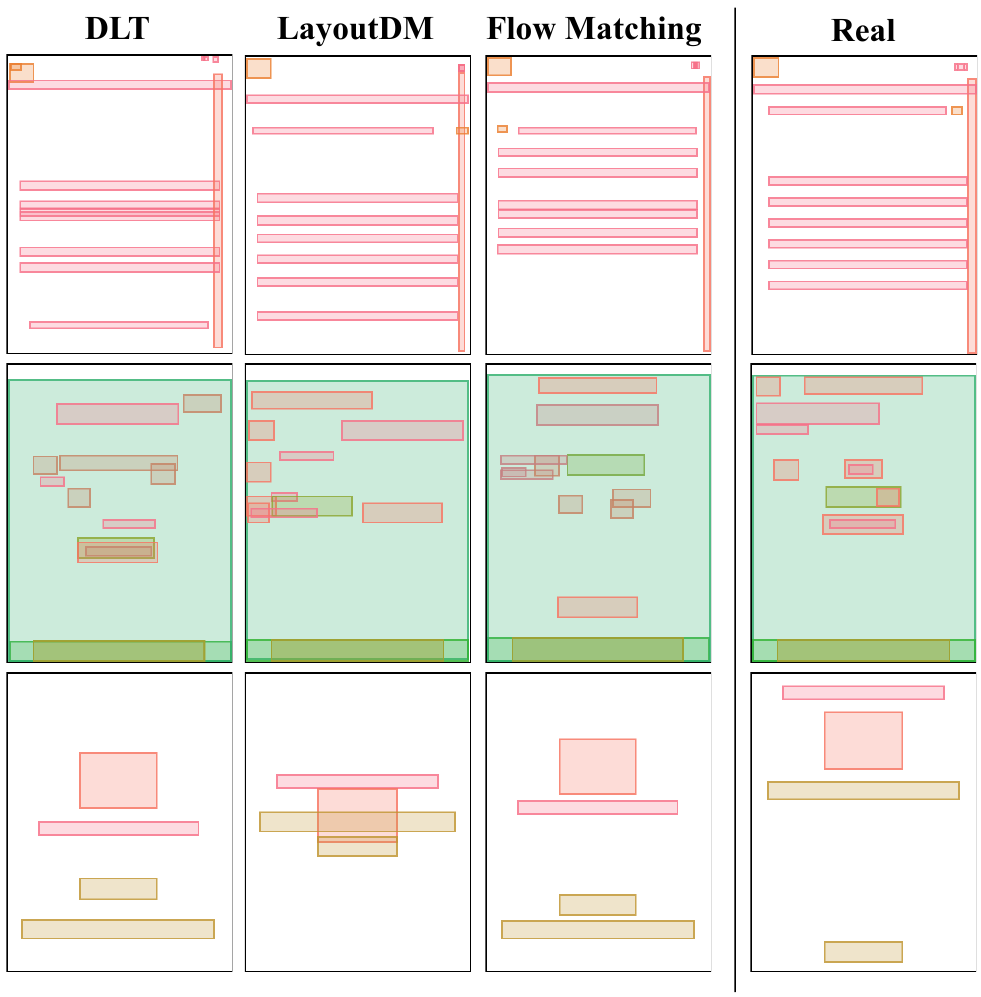}
        \caption{\textbf{Examples for \tasktypesize}}
    \end{subfigure}
    \caption{\textbf{Conditional Generation Examples}}
    \label{fig:qual_cond}
\end{figure}

\textbf{Qualitative Evaluation.} We provide some qualitative examples in \cref{fig:qual_uncond} for the unconditional \cref{fig:qual_cond} as well as for the conditional generation task on the RICO dataset. More samples for the other tasks and PubLayNet can be found in the supplementary material. Overall, LayoutFlow shows a strong performance across all tasks, producing visually pleasing results that resemble real layouts.

\textbf{Limitations.} We discuss the limitations of our proposed methods in the supplementary material.

\subsection{Ablation Study}
In order to justify our design choices, we perform an ablation study on unconditional layout generation using the RICO dataset.

\textbf{Diffusion vs. Flow.}
We show that employing Flow Matching as the generative model is essential by also training the model used for LayoutFlow and only substituting Flow Matching with diffusion. We test two methods for sampling from the diffusion-based model, i.e., DDPM and DDIM. As shown in \cref{tab:flow_diff}, neither approach is able to reach the same performance as LayoutFlow, and both seem to be particularly lacking in terms of \textit{Overlap}. Interestingly, DDIM outperforms DDPM by a large margin, which might be related to DDIM sampling based on an ODE instead of an SDE.

\textbf{Regularization Loss.} We validate the efficacy of our regularization loss proposed in \cref{eq:loss} and the choice of $\lambda$ in \cref{tab:loss_func}. It can be clearly observed that a stronger regularization using the $L1$-loss provides better alignment but increases the \textit{FID}. Therefore, we considered the trade-off across all metrics and chose $\lambda=0.2$ for LayoutFlow. 

\begin{table}[t]
    \begin{minipage}[b]{0.46\textwidth}
        \caption{\textbf{Ablation study on the choice of training approaches using the same architecture}}
        \label{tab:flow_diff}
        \centering
        \begin{tabular}{lrrrrr}
            \toprule 
            Approach & \quad & \abbfid & \abbalignment & \abboverlap & \abbmiou \\
            \midrule
            Diff.(DDPM) & \quad     & 34.89             & \bftab{0.128} & 0.335           & 0.116\\
            Diff.(DDIM) & \quad     & 3.18              & 0.181         & 0.561           & 0.548\\
            Flow        & \quad     & \bftab{2.37}      & 0.150         & \bftab{0.498}   & \bftab{0.570}\\
            \bottomrule
        \end{tabular}
    \end{minipage}
    \hspace{0.25cm}
    \begin{minipage}[b]{0.46\textwidth}
        \caption{\textbf{Ablation study on different loss functions}}
        \label{tab:loss_func}
        \begin{tabular}{lrrrr}
            \toprule
            Loss& \abbfid & \abbalignment & \abboverlap & \abbmiou \\
            \midrule
            $\mathcal{L}_{\mathrm{CFM}}$                        & \np{2.27}     &  0.194        & 0.507         & 0.570\\
            $\mathcal{L}_{\mathrm{CFM}} + 0.1 \mathcal{L}_1$    & \bftab{2.20}  & 0.155         & \bftab{0.495} & 0.559\\
            $\mathcal{L}_{\mathrm{CFM}} + 0.2 \mathcal{L}_1$    & \np{2.37}     & 0.150         & 0.498         & 0.570\\
            $\mathcal{L}_{\mathrm{CFM}} + 0.3 \mathcal{L}_1$    & \np{2.40}     & 0.144         & 0.498         & 0.581\\
            $\mathcal{L}_{\mathrm{CFM}} + 0.4 \mathcal{L}_1$    & \np{2.60}     & \bftab{0.134} & 0.498         & 0.579\\
            $L_1$-loss only                                     & \np{26.64}    &  0.159        & 0.625         & \bftab{0.586}\\
            \bottomrule
        \end{tabular}
    \end{minipage}
\end{table}

\textbf{Initial Distribution.} Since flow-based models allow for arbitrary initial distributions, we explore two alternatives. First, we try a uniform distribution to ensure all initial samples are placed on the canvas. In addition, as Analog Bits have a smaller sample space, we tested using a Gaussian for the geometrical elements and a uniform distribution for the Analog Bits. Altogether, the results in \cref{tab:init_dist} indicate that a Gaussian initial distribution provides the best overall performance measured by the \textit{FID} score, whereas a uniform distribution provides stronger results in terms of geometrical metrics, such as \textit{Alignment} and \textit{Overlap}.   

\begin{table}[t]
    \begin{minipage}[b]{0.46\textwidth}
        \centering
        \caption{\textbf{Ablation study on different initial distributions}}
        \begin{tabular}{lrrrr}
            \toprule
            Distribution & \abbfid & \abbalignment & \abboverlap & \abbmiou \\
            \midrule
            Gaussian & \bftab{2.37} & 0.150         & 0.498         & 0.570 \\
            Uniform  & 2.61         & \bftab{0.115} & \bftab{0.481} & \bftab{0.584}\\
            Mixture  & 2.52         & 0.160         & 0.494         & 0.562 \\
            \bottomrule
        \end{tabular}
        \label{tab:init_dist}
    \end{minipage}
    \hspace{0.25cm}
    \begin{minipage}[b]{0.46\textwidth}
        \centering
        \caption{\textbf{\textbf{Ablation study on different training trajectories}}}
        \begin{tabular}{lrrrr}
            \toprule
            Trajectory & \abbfid & \abbalignment & \abboverlap & \abbmiou \\
            \midrule
            Linear      & 2.37          & \bftab{0.150} & 0.498         & \bftab{0.570}\\
            Sine/Cosine & \bftab{2.33}  & 0.172         & 0.533         & 0.557\\
            Sine        & 2.48          & 0.152         & \bftab{0.480} & 0.565\\
            \bottomrule
        \end{tabular}
        \label{tab:train_traj}
    \end{minipage}
\end{table}

\textbf{Training Trajectories.} As long as the conditional vector field fulfills the criteria presented in \cref{sec:prelim_flow}, it can be freely chosen for training. In addition to the linear training trajectory proposed in~\cite{lipman2023flow, tong2023improving}, we also explore the sine/cosine interpolation introduced in~\cite{albergo2023building} and a sine-based interpolation, as shown in \cref{tab:train_traj}. The choice of training trajectories only slightly affects the performance.

\section{Conclusion}
In this paper, we explored the application of Flow Matching for layout generation and, as a result, proposed LayoutFlow, a flow-based model that is able to handle various layout generation tasks. Our model is able to significantly speed up inference compared to other diffusion-based models while providing state-of-the-art performance. Even though we tested a diverse set of design options, there still remains room for further exploration into applying Flow Matching to layout generation, for example, different training trajectories, conditioning methods, or initial distributions. Furthermore, LayoutFlow might also be extended to generate layouts considering the content in the elements, similar to related work~\cite{xu2023unsupervised, hsu2023posterlayout}. Overall, we demonstrate that Flow Matching provides a highly flexible and powerful tool for layout generation that offers a natural geometrical interpretation.   

\section*{Acknowledgments}
This work was supported by JSPS KAKENHI Grant Number JP23K28139.

\bibliographystyle{splncs04}
\bibliography{egbib}

\appendix
\clearpage

\begin{center}
	\vspace*{0.5cm}
	\Large
	\textbf{Supplementary Material}
	\vspace{0.75cm}
\end{center}

In this document, we provide additional explanations and evaluations to supplement the contents of the main paper. Specifically, we present further discussions on training trajectories, conditioning, and the limitations of our method, as well as additional quantitative and qualitative results. 

\section{Training Trajectories}
The conditional flow, or training trajectory, $\phi_t(\mathbf{x}) = \mathbf{x}_t$ can be chosen freely as long as the conditional vector field follows the continuity equation and $\phi_0(\mathbf{x}) = \mathbf{x}_0$ and $\phi_1(\mathbf{x}) = \mathbf{x}_1$. While LayoutFlow is trained using a linear trajectory following~\cite{lipman2023flow, tong2023improving}, we also tested the sine/cosine interpolation introduced in~\cite{albergo2023building} and a sine-based interpolation. More details are shown in \cref{tab:train_traj2}. Our main motivation for the sine-based interpolation was to preserve the direction of the conditional vector field but apply a different time schedule to the trajectory. In the linear scenario, the conditional vector field remains independent of the time and always points towards $x_1$. On the other hand, the sine/cosine trajectory changes its direction over time, only gradually leading to $x_1$ in a circular trajectory. The sine-based interpolation maintains the direction of the linear trajectory but weighs it depending on the time. At an early time of the trajectory, the magnitude of the direction is large, whereas towards the end, the derivative starts to slow down. In analogy to time importance sampling in diffusion models~\cite{levi2023dlt}, we hypothesized that this might help focus on the later stages of the flow process during training, where more detailed features become important. However, based on the paper's ablation results, the cosine-based interpolation does not seem to affect the performance. Nonetheless, this scheduling property of conditional fields might be further explored with other functions of the general form:
\begin{align}
    v_t = \kappa(t) (\mathbf{x}_1 - \mathbf{x}_0),
\end{align}
while still fulfilling the conditions mentioned above. In the cosine case of $\kappa(t) = \cos(\frac{pi}{2}t)$, the schedule might have been too close to the linear schedule to make a noticeable change, and different choices might provide better results. 

\begin{table}[t]
    \centering
    \caption{\textbf{Overview of different training trajectories and conditional vector fields.} Note that the conditional vector field is the time-derivative of the training trajectory.}
    \begin{tabular}{lrrr}
        \toprule
        Name        &   Training Trajectory $\mathbf{x}_t$                                                  &  \quad & Conditional Vector Field $v_t$ \\
        \midrule
        Linear      &   $(1-t) \mathbf{x}_0 + t \mathbf{x}_1$                                               &   & $\mathbf{x}_1 - \mathbf{x}_0$  \\
        Sine/Cosine &   $\cos{(\frac{\pi}{2}t)} \mathbf{x}_0 + \sin{(\frac{\pi}{2}t)} \mathbf{x}_1$         &   & $\cos{(\frac{\pi}{2}t)} \mathbf{x}_1-\sin{(\frac{\pi}{2}t)} \mathbf{x}_0$ \\
        Sine        &   $ (1-\sin{(\frac{\pi}{2}t})) \mathbf{x}_0 + \sin{(\frac{\pi}{2}t)} \mathbf{x}_1$    &   & $\cos{(\frac{\pi}{2}t)} (\mathbf{x}_1 - \mathbf{x}_0)$ \\
        \bottomrule
    \end{tabular}
    \label{tab:train_traj2}
\end{table} 

\section{Inference Speed Analysis}
Since previous layout generation approaches rely on different architectures, we provide some insights on how the architectural components, \ie, the model size measured by the numbers of trainable parameters and the number of tokens needed to represent a single layout, influence the inference speed in Table \ref{tab:speed_up}. In addition, we also report the number of steps each method requires to generate a layout that is tied to the choice of the generative model. Overall, it can be seen that the speed-up for LayoutFlow can be attributed to architectural improvements as well as changing from a diffusion model to a flow-base model due to the reduced number of generation steps required.   

\begin{table}[h]
    \centering
    \caption{\textbf{Overview of different factors contributing to inference speed}}
    \begin{tabular}{lrrrr}
        \toprule 
                            & Model Size  & Token per Layout  & Generation Steps  &  Inference Speed \\
        \midrule
        LayoutDiffusion     & 86M               & $>$100            & 160               & 1600.00ms\\
        LayoutDM            & 12.4M             & 100               & 100               & 16.60ms\\
        DLT                 & 9.0M              & 20                & 100               & 3.50ms\\
        LayoutFlow (ours)   & 12.7M             & 20                & 50                & \bftab{1.75ms}\\
        \bottomrule
    \end{tabular}
    \label{tab:speed_up}
\end{table}

\section{Quality Speed Trade-off}
In \cref{fig:performance_time}, we compare the number of function evaluations of our model, corresponding to the number of steps taken to solve the ODE, to the performance on the RICO dataset. While only 5 function evaluations result in a bad score of an \textit{FID} of roughly 50, just three more evaluations can bring down the \textit{FID} to 13. The improvement observed by increasing the number of steps plateaus at around 40. In comparison, diffusion models typically require at least 100 steps to perform well. Overall, LayoutFlow offers a strong quality-speed trade-off, where slightly reducing the number of steps, \eg, from 30 to 20, only has a small impact on performance. 
\begin{figure}
    \centering
    \begin{tikzpicture}
    \begin{axis}[
        xlabel={Number of function evaluations},
        ylabel={FID score},
        y label style={yshift=-14pt},
        xmode=log,
        ymode=log,
        xmin=4, xmax=200,
        ymin=1.5, ymax=60,
        xtick={5, 8, 10, 20, 30, 40, 50, 100, 150.01},
        ytick={2, 3, 5, 10, 20, 50},
        legend pos=north east,
        legend style={cells={anchor=west}}, 
        xmajorgrids=true,
        xminorgrids=true,
        ymajorgrids=true,
        yminorgrids=true,
        grid style=dashed,
        xticklabel ={
            \pgfmathparse{exp(\tick)}
            \pgfmathprintnumber{\pgfmathresult}
        }, 
        yticklabel={
            \pgfmathparse{exp(\tick)}
            \pgfmathprintnumber{\pgfmathresult}
        }, 
    ]   
    \addplot[
        color=blue,
        mark=diamond*,
        ]
        coordinates {
        (5, 52.427)
        (8, 12.953)
        (10, 7.473)
        (20, 3.012)
        (30, 2.480)
        (40, 2.330)
        (50, 2.330)
        (100, 2.368)
        (150, 2.335)
        };
    \addplot[name path=std_up, color=red!70]coordinates{
        (5, 51.989)
        (8, 12.470)
        (10, 7.177)
        (20, 2.921)
        (30, 2.369)
        (40, 2.257)
        (50, 2.241)
        (100, 2.249)
        (150, 2.239)
    };
    \addplot[name path=std_down, color=red!70]coordinates{
        (5, 52.864)
        (8, 13.437)
        (10, 7.769)
        (20, 3.104)
        (30, 2.591)
        (40, 2.402)
        (50, 2.419)
        (100, 2.486)
        (150, 2.431)
    };
    \addplot[red!50,fill opacity=0.5] fill between[of=std_down and std_up];

    \end{axis}
\end{tikzpicture}
    \caption{\textbf{Quality-Speed Trade-Off.} We investigate the relationship between the step size employed by the Euler method, which translates to the number of function evaluations, and the \textit{FID} score using LayoutFlow trained on the RICO dataset.}
    \label{fig:performance_time}
\end{figure}
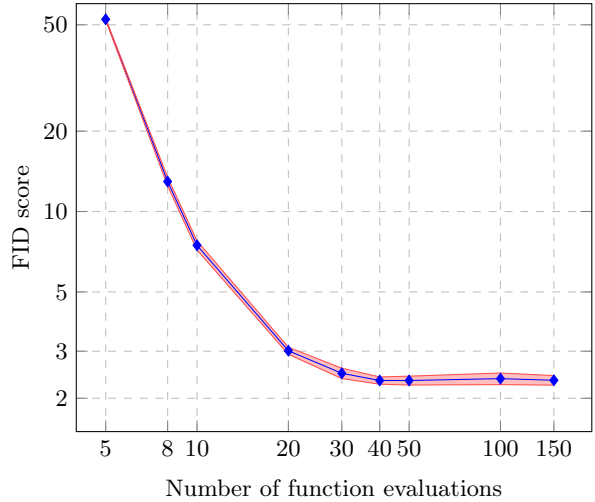    

\section{Conditioning Analysis}
There are several ways to introduce conditions to a generative task. For layout generation, given conditions can differ significantly depending on the task. For LayoutFlow, we employ a masking strategy during training that indicates which part of the input serves as a condition, similar to
~\cite{chen2024towards, hui2023unifying, levi2023dlt}. Intuitively, the masking mechanism can be considered a switch that allows training multiple flow-based task-specific models with a single shared architecture in just one model. Alternatively, it is possible only to train an unconditional model and impose the conditions during sampling through a guidance strategy as done in ~\cite{inoue2023layoutdm, zhang_layoutdiffusion}. This strategy only requires training the model on the unconditional task and inserting the condition during inference, effectively altering the trajectory.

To condition the trajectory during inference, we employ a mask $\mathbf{m}$ that is zero for the condition dimension, \ie, all ones in the case of unconditional generation. Essentially, we use the mask only to update the direction of the condition dimension to $\mathbf{u'}$ and can be described as  

\[
    \mathbf{u'}_{\frac{k+1}{T}} = \mathbf{m} \odot \mathbf{u}_{\theta}( \mathbf{x}_{\frac{k}{T}} + {\frac{1}{T}} \mathbf{u'}_{\frac{k}{T}}) + (1-\mathbf{m}) \odot ( \mathbf{x}_c - \mathbf{x}_{\frac{k}{T}} ),
\]
where $\mathbf{x}_c$ denotes the conditions and $\odot$ represents the Hadamard product. For the conditional task, we update the direction only while $k<0.8T$ as we find it to work better empirically.  
\Cref{tab:cond_abl} compares both conditioning methods and illustrates that a training-based approach significantly outperforms conditioning during inference. While we tried several other methods to introduce the conditions during inference, there might be more sophisticated conditioning methods that perform better and might be a potential future research direction.

\begin{table}[t]
    \centering
    \caption{\textbf{Quantitative comparison of training LayoutFlow on multiple tasks using a masking strategy and inserting conditions to an unconditional LayoutFlow model by modifying the trajectory during inference.} 
    The best results are highlighted in \textbf{bold}.}
    \begin{tabular}{llrrrr}
        \toprule
        & &  \multicolumn{4}{c}{RICO} \\
        Task & Condition &  \abbfid & \abbalignment & \abboverlap & \abbmiou  \\
        \midrule
        \multirow{2}{*}{\taskugen}& None &  \bftab{2.28}&  0.155&  0.511&   \bftab{0.610} \\
        & Mask&  2.37&  \bftab{0.150}&  \bftab{0.498}&   0.570\\ 
        \midrule
        \multirow{2}{*}{\tasktype}& Trajectory&  33.38&  0.202&  0.635&   0.160\\
        & Mask&  \bftab{1.48}&  \bftab{0.176}&  \bftab{0.517}&   \bftab{0.322}\\ 
        \midrule
       \multirow{2}{*}{\tasktypesize}& Trajectory&  174.66&  0.862&  0.728&   0.181\\
        & Mask&  \bftab{1.03}&  \bftab{0.283}&  \bftab{0.523}&   \bftab{0.470}\\ 
        \midrule
       \multirow{2}{*}{\taskcompletion}& Trajectory&  7.96&  0.193&  0.520&  0.657 \\
        & Mask&  \bftab{1.51}&  \bftab{0.150}&  \bftab{0.474}&   \bftab{0.741}\\ 
        \midrule
        \bftab{Validation Dataset}&  & 2.10& 0.093& 0.466&  0.658\\
        \bottomrule
    \end{tabular}
    \label{tab:cond_abl}
\end{table}

\section{Results Using Different Data Split}
Due to the different data splits used by previous methods, comparing the numbers reported by other papers has become difficult. The FID values depend on a separately trained network, making comparing even more difficult as the same weights and network are required. Our experiments used the same settings as in~\cite{zhang_layoutdiffusion, jiang2023layoutformer++}. As a result, we had to retrain LayoutDM~\cite{inoue2023layoutdm} and DLT~\cite{levi2023dlt} with that specific data split. To ensure that LayoutFlow works robustly across different dataset settings, we also trained a model on the data split used in LayoutDM~\cite{chai2023layoutdm}, which includes up to 25 elements per layout as opposed to only 20 elements in the other setting. Additionally, this allows us to compare with a concurrent work called LACE~\cite{chen2024towards}, which uses a continuous diffusion model.       

\begin{table}[t]
    \centering
    \caption{\textbf{Quantitative results on a different dataset split~\cite{inoue2023layoutdm} for the PubLayNet dataset.} 
    The best result is highlighted in \textbf{bold}. The $\rightarrow$ symbol indicates best results are the ones closest to the validation data.}
    \begin{tabular}{llrrrr}
        \toprule
        & &  \multicolumn{4}{c}{PubLayNet}\\
        Task & Model &  \abbfid & \abbalignment & \abboverlap & \abbmiou \\
        \midrule
        \multirow{3}{*}{\taskugen} 
        & LACE& 8.45& 0.141& 0.075& -\\
        & LayoutDM                          &  13.90&  0.195&  0.134&  -\\
        & LayoutFlow (ours)                     &  \bftab{7.48}&  \bftab{0.066}& \bftab{0.015}& 0.423\\ 
        \midrule
        \multirow{3}{*}{\tasktype}
        & LACE &  5.14&  0.046&  0.018&  \bftab{0.383}\\
        & LayoutDM                  &  7.95&  0.106&  0.164& 0.310\\
        & LayoutFlow (ours)         & \bftab{3.58}& \bftab{0.046}& \bftab{0.018}&  0.349\\ 
        \midrule
        \multirow{3}{*}{\tasktypesize}
        & LACE &  2.66&  0.061&  \bftab{0.034}&  0.418\\
        & LayoutDM&  4.25&  0.119&  0.189&  0.381\\
        & LayoutFlow (ours)       &  \bftab{0.80}& \bftab{0.052}&  0.036& \bftab{0.443}\\ 
        \midrule
        \textbf{Validation Data} && 6.25 & 0.021& 0.003&  0.438\\
        \bottomrule
    \end{tabular}
    \label{tab:quant_results}
\end{table}

\section{Completion Task}
The completion task describes generating a complete layout given an incomplete layout. In this scenario, the given incomplete layout acts as a strong condition. There are different ways to define the completion task, \eg, Inoue \etal assume that up to 20\% of the layout is already given. While this scenario is closer to unconditional generation as it still offers a high degree of flexibility, we also looked into completing almost finished layouts containing more than 80\% of their elements. To show that LayoutFlow can handle either scenario, we seperately trained models with the respective conditional mask and present the results in \cref{tab:compl_add} compared to LayoutDM and show the qualitative results in \cref{sec:more_qual}.  

\begin{table}[t]
    \centering
    \caption{\textbf{Quantitative results for the completion task with different percentages of the missing elements on the RICO and PubLayNet datasets.} 
    The best results are highlighted in \textbf{bold}. Models marked with * have been retrained.}
    \resizebox{\columnwidth}{!}{
    \begin{tabular}{llrrrrrrrrr}
        \toprule
        & &  \multicolumn{4}{c}{RICO} &&  \multicolumn{4}{c}{PubLayNet}\\
        Task & Model &  \abbfid & \abbalignment & \abboverlap & \abbmiou & $\quad$ & \abbfid & \abbalignment & \abboverlap & \abbmiou \\
        \midrule
        \multirow{2}{*}{Completion (20\%)}& LayoutDM*&  6.80&  \bftab{0.054}&  0.630&   0.678&&  25.02&  0.169&  0.107& 0.678\\
        & LayoutFlow (ours)                     &  \bftab{1.51}&  0.150&  \bftab{0.474}&   \bftab{0.741}&&  \bftab{1.10}&  \bftab{0.054}&  \bftab{0.127}& \bftab{0.746}\\ 
        \midrule
       \multirow{2}{*}{Completion (80\%)}& LayoutDM*&  5.21&  \bftab{0.094}&  0.658&   0.574&&  28.73&  0.223&  0.146& 0.385\\
        & LayoutFlow (ours)                     &  \bftab{3.59}&  0.182&  \bftab{0.605}&   \bftab{0.628}&&  \bftab{4.02}&  \bftab{0.050}&  \bftab{0.024}& \bftab{0.445}\\ 
        \midrule
        \textbf{Validation Dataset}&  & 2.10& 0.093& 0.466&  0.658&& 8.10& 0.022& 0.003&0.434\\
        \bottomrule
    \end{tabular}
    }
    \label{tab:compl_add}
\end{table}

\section{Limitations and Broader Impact}
While our work aims to improve the state of layout generation further and help make design tasks easier, there are also some potential negative effects. Easier and better layout generation might be misused to build spam websites or large amounts of content or potentially be used for scams, particularly phishing attempts. The potential harm does not come from enabling such actions but derives from the scale that automation allows. Nonetheless, we believe the democratization of technologies like ours outweighs these risks.    

Our proposed model, LayoutFlow, has shown remarkable results. However, some areas can still be improved in the future. Alignment remains one of the largest challenges, especially for models working in the continuous data space. While LayoutFlow utilizes an L1 regularization loss to improve the alignment, the issue is not fully resolved. For the future, it is crucial to find a better loss function that can reflect the perceptual error between layouts rather than relying too heavily on metrics that do not necessarily correlate with important design aspects such as alignment.

\section{More Qualitative Results}
\label{sec:more_qual}
Due to limited space in the main paper, we present more qualitative results for each task and dataset in this section.

\begin{center}
    
\begin{table}
    \centering
    \begin{tabular}{cccc}
        
        DLT & LayoutDM & LayoutDiffusion & LayoutFlow \\
        \includegraphics[width=0.2\textwidth]{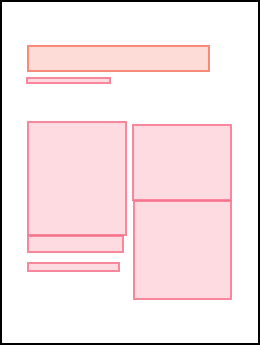} & \includegraphics[width=0.2\textwidth]{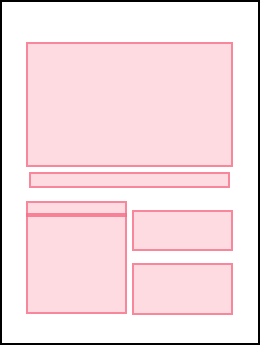} & \includegraphics[width=0.2\textwidth]{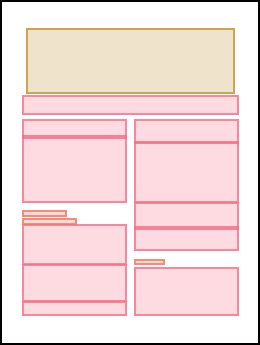} & \includegraphics[width=0.2\textwidth]{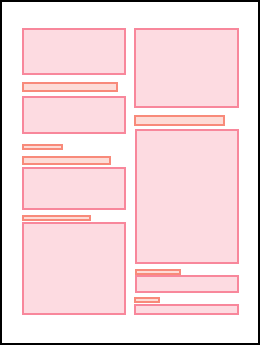} \\
        \includegraphics[width=0.2\textwidth]{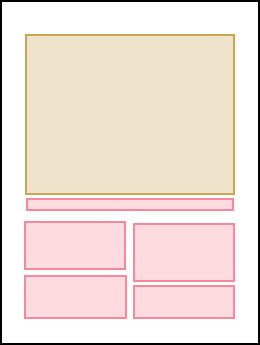} & \includegraphics[width=0.2\textwidth]{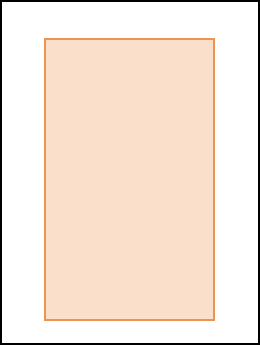} & \includegraphics[width=0.2\textwidth]{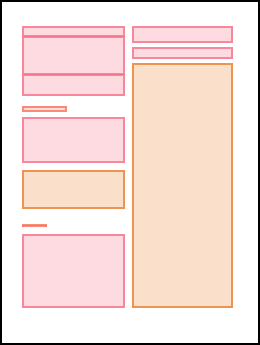} & \includegraphics[width=0.2\textwidth]{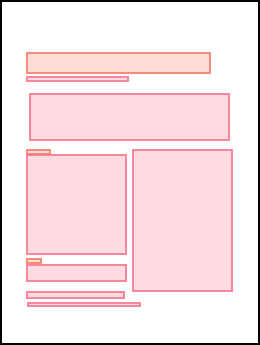} \\
        \includegraphics[width=0.2\textwidth]{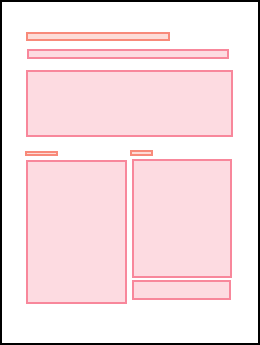} & \includegraphics[width=0.2\textwidth]{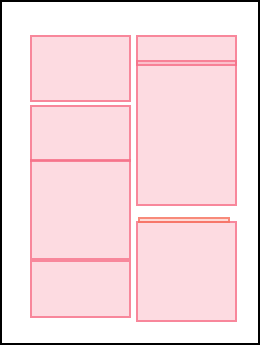} & \includegraphics[width=0.2\textwidth]{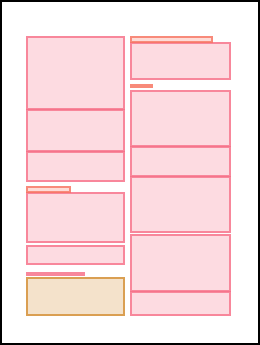} & \includegraphics[width=0.2\textwidth]{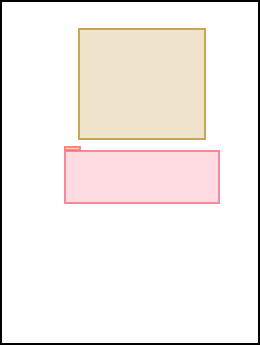} \\
        \includegraphics[width=0.2\textwidth]{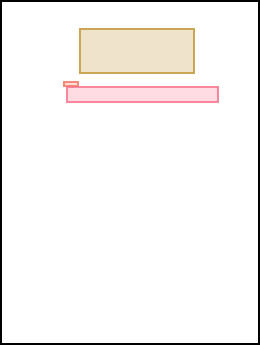} & \includegraphics[width=0.2\textwidth]{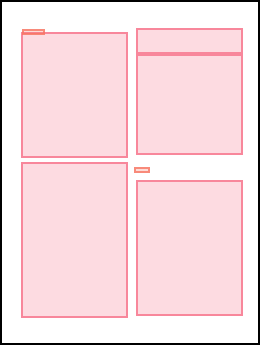} & \includegraphics[width=0.2\textwidth]{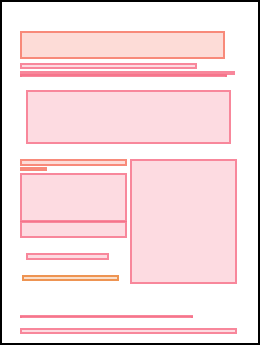} & \includegraphics[width=0.2\textwidth]{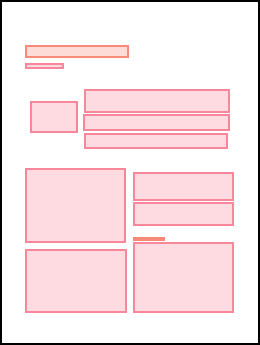} \\
        \includegraphics[width=0.2\textwidth]{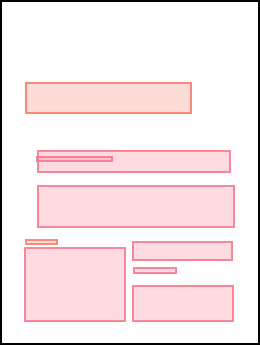} & \includegraphics[width=0.2\textwidth]{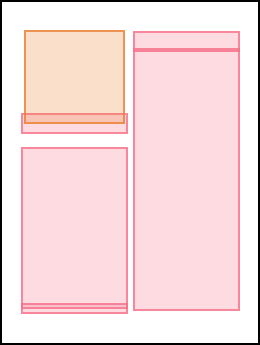} & \includegraphics[width=0.2\textwidth]{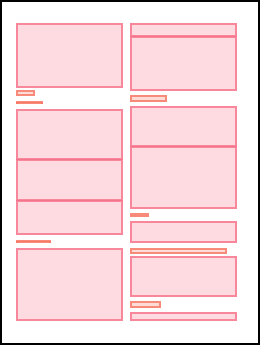} & \includegraphics[width=0.2\textwidth]{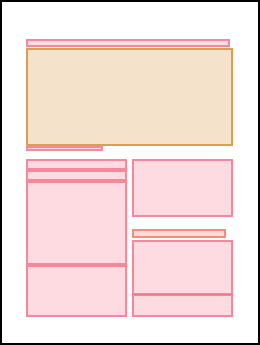} \\

    \end{tabular}
    \caption{Unconditional Generation on PubLayNet}
    
\end{table}

\begin{table}
    \centering
    \begin{tabular}{cc|ccccccc|cc}
        
        Input & \quad & \quad &  DLT & LayoutDM & LayoutDiff. & LayoutFlow & \quad & \quad & & Ground Truth \\
        \includegraphics[width=0.12\textwidth]{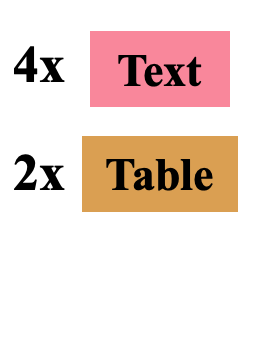} 
        & \quad & \quad &
        \includegraphics[width=0.12\textwidth]{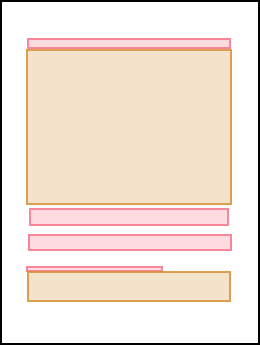} & 
        \includegraphics[width=0.12\textwidth]{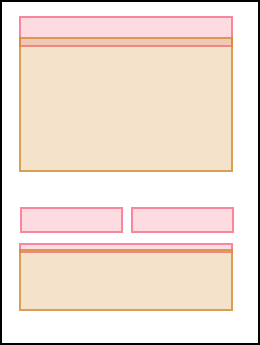} & 
        \includegraphics[width=0.12\textwidth]{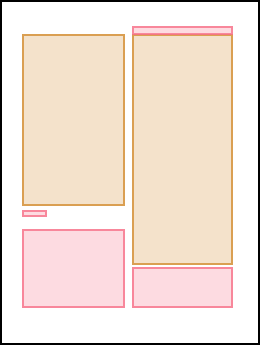} &
        \includegraphics[width=0.12\textwidth]{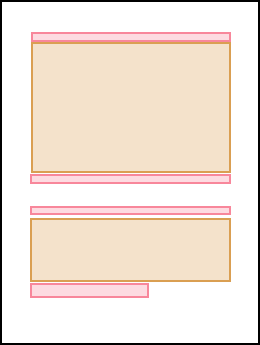} & 
        & \quad & \quad &
        \includegraphics[width=0.12\textwidth]{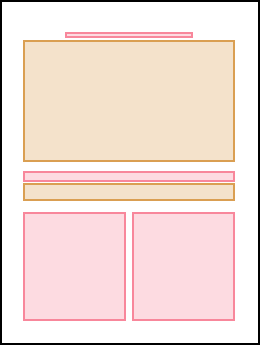} \\
        \includegraphics[width=0.12\textwidth]{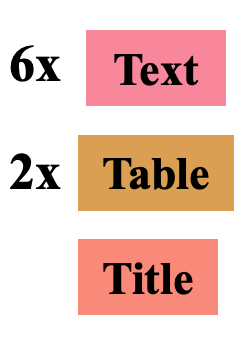} 
        & \quad & \quad &
        \includegraphics[width=0.12\textwidth]{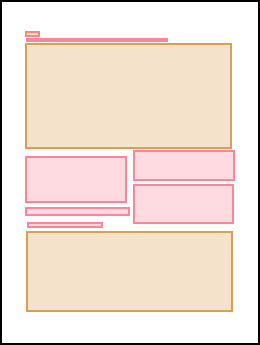} & 
        \includegraphics[width=0.12\textwidth]{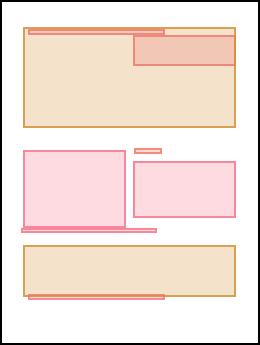} & 
        \includegraphics[width=0.12\textwidth]{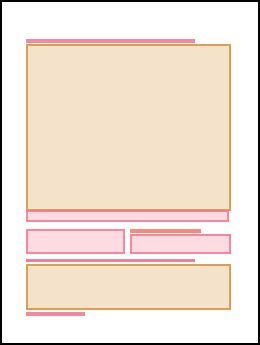} &
        \includegraphics[width=0.12\textwidth]{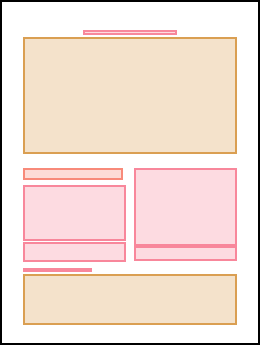} & 
        & \quad & \quad &
        \includegraphics[width=0.12\textwidth]{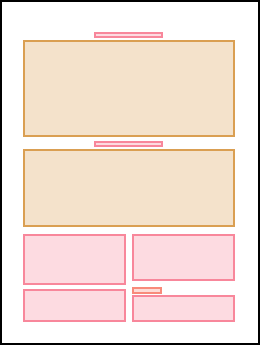} \\
        \includegraphics[width=0.12\textwidth]{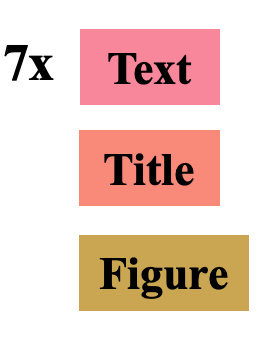} 
        & \quad & \quad &
        \includegraphics[width=0.12\textwidth]{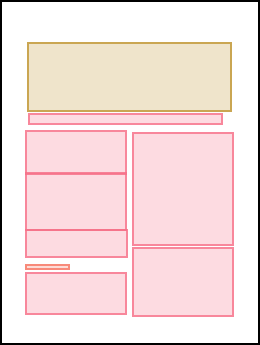} & 
        \includegraphics[width=0.12\textwidth]{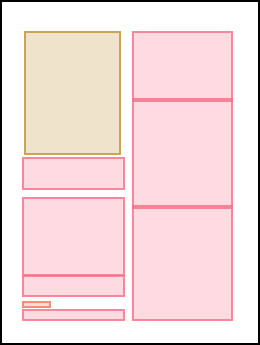} & 
        \includegraphics[width=0.12\textwidth]{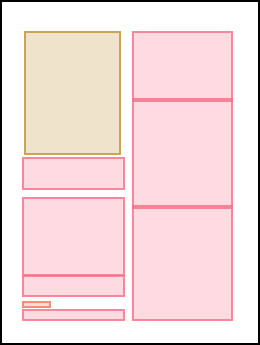} &
        \includegraphics[width=0.12\textwidth]{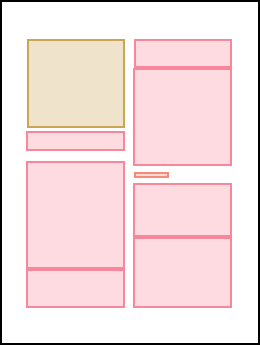} & 
        & \quad & \quad &
        \includegraphics[width=0.12\textwidth]{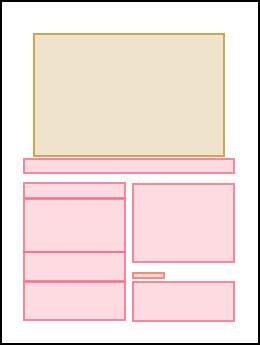} \\
        \includegraphics[width=0.12\textwidth]{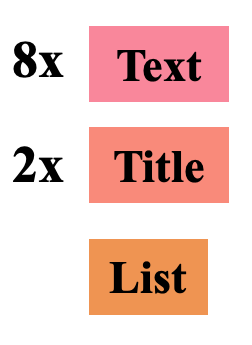} 
        & \quad & \quad &
        \includegraphics[width=0.12\textwidth]{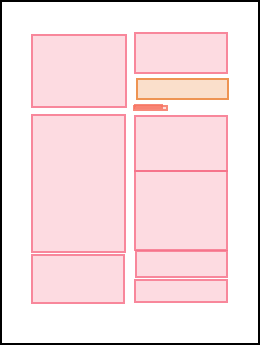} & 
        \includegraphics[width=0.12\textwidth]{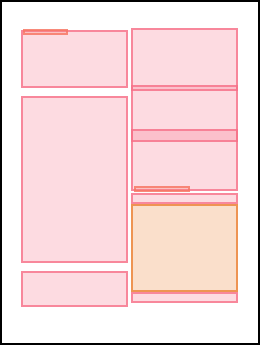} & 
        \includegraphics[width=0.12\textwidth]{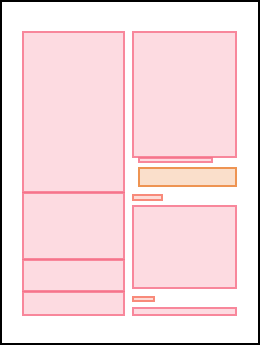} &
        \includegraphics[width=0.12\textwidth]{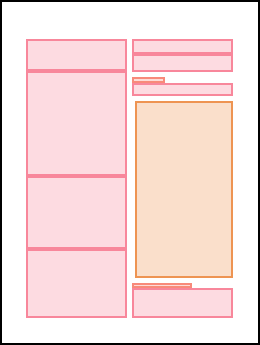} & 
        & \quad & \quad &
        \includegraphics[width=0.12\textwidth]{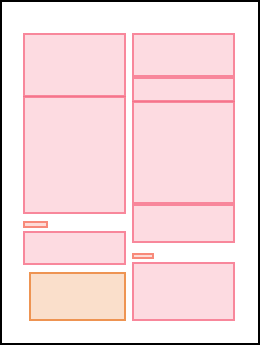} \\
        \includegraphics[width=0.12\textwidth]{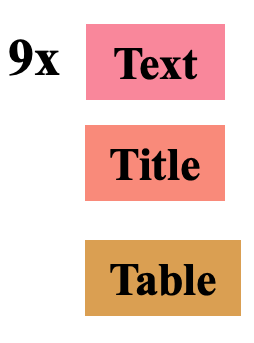} 
        & \quad & \quad &
        \includegraphics[width=0.12\textwidth]{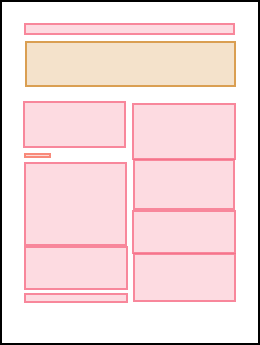} & 
        \includegraphics[width=0.12\textwidth]{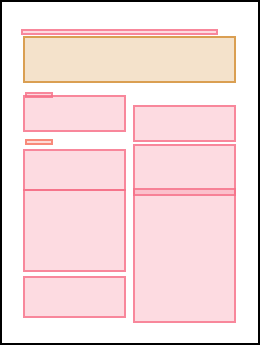} & 
        \includegraphics[width=0.12\textwidth]{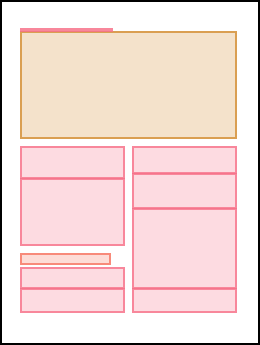} &
        \includegraphics[width=0.12\textwidth]{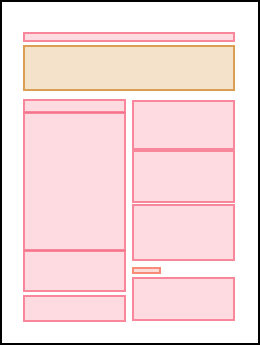} & 
        & \quad & \quad &
        \includegraphics[width=0.12\textwidth]{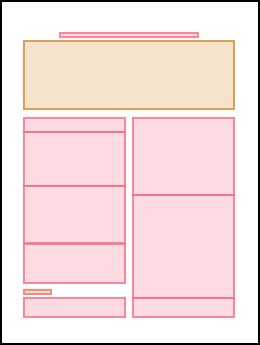} \\

    \end{tabular}
    \caption{Category Conditional Generation on PubLayNet}
    
\end{table}

\begin{table}
    \centering
    \begin{tabular}{cccc|cc}
        
        DLT & LayoutDM & LayoutFlow & \quad & \quad & Ground Truth \\
        \includegraphics[width=0.2\textwidth]{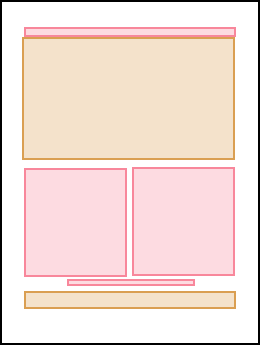} & \includegraphics[width=0.2\textwidth]{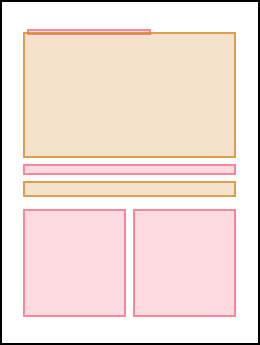} & \includegraphics[width=0.2\textwidth]{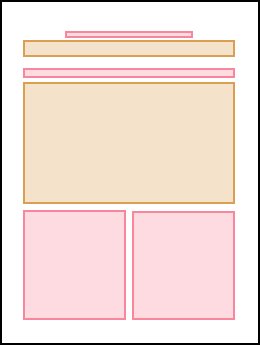} & \quad & \quad & \includegraphics[width=0.2\textwidth]{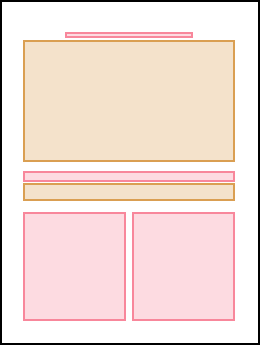} \\
        \includegraphics[width=0.2\textwidth]{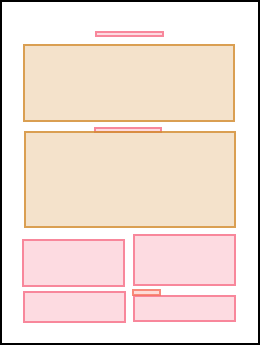} & \includegraphics[width=0.2\textwidth]{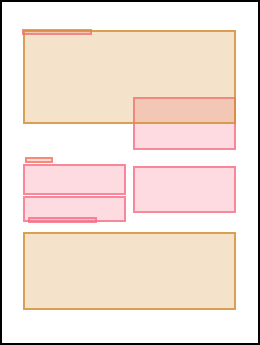} & \includegraphics[width=0.2\textwidth]{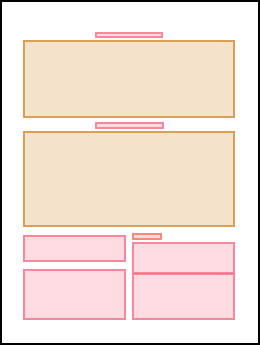} & \quad & \quad & \includegraphics[width=0.2\textwidth]{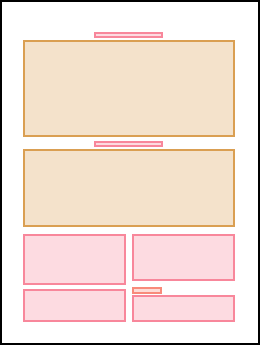} \\
        \includegraphics[width=0.2\textwidth]{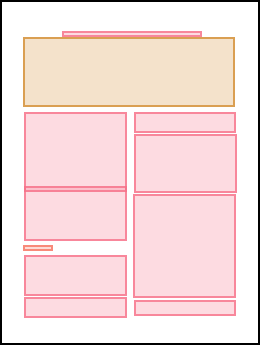} & \includegraphics[width=0.2\textwidth]{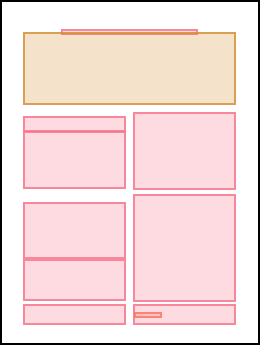} & \includegraphics[width=0.2\textwidth]{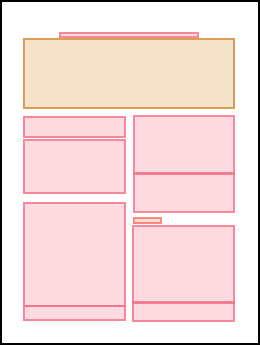} & \quad & \quad & \includegraphics[width=0.2\textwidth]{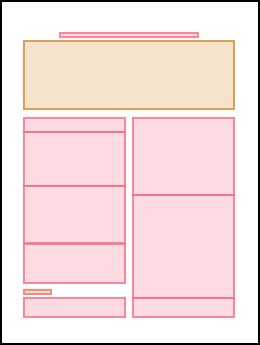} \\
        \includegraphics[width=0.2\textwidth]{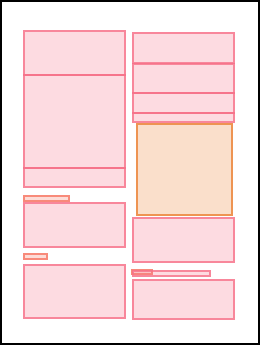} & \includegraphics[width=0.2\textwidth]{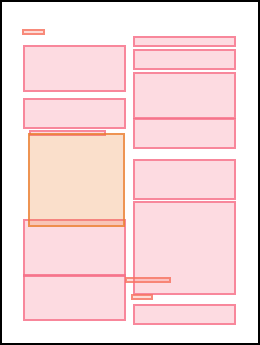} & \includegraphics[width=0.2\textwidth]{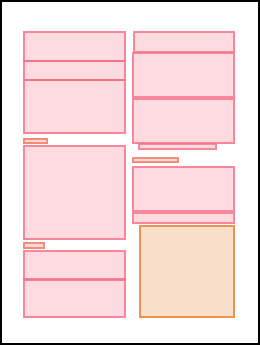} & \quad & \quad & \includegraphics[width=0.2\textwidth]{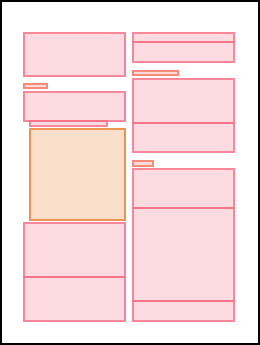} \\
        \includegraphics[width=0.2\textwidth]{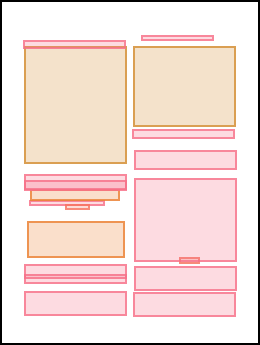} & \includegraphics[width=0.2\textwidth]{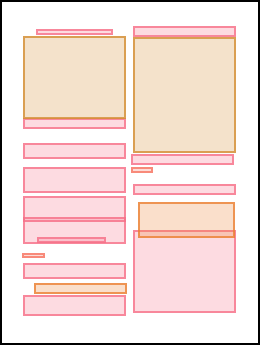} & \includegraphics[width=0.2\textwidth]{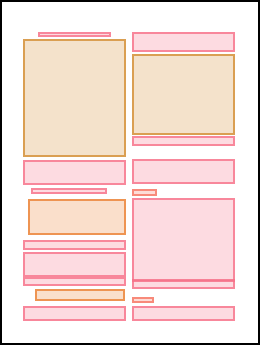} & \quad & \quad & \includegraphics[width=0.2\textwidth]{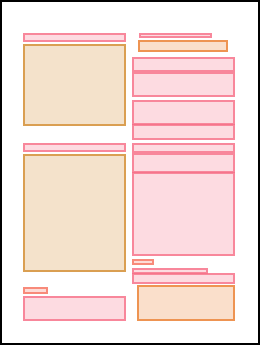} \\

    \end{tabular}
    \caption{Size Conditional Generation on PubLayNet}
    
\end{table}

\begin{table}
    \centering
    \begin{tabular}{cc|cccc|cc}
        
        Input & \quad & \quad &  LayoutDM & LayoutFlow & \quad & \quad & Ground Truth \\
        \includegraphics[width=0.2\textwidth]{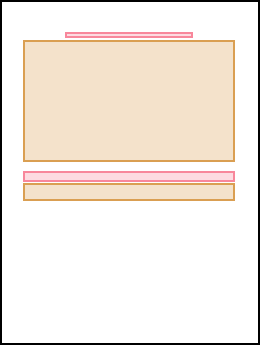} & \quad & \quad &\includegraphics[width=0.2\textwidth]{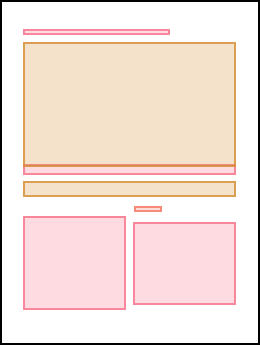} & 
        \includegraphics[width=0.2\textwidth]{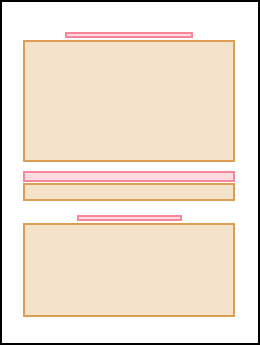} & \quad & \quad &\includegraphics[width=0.2\textwidth]{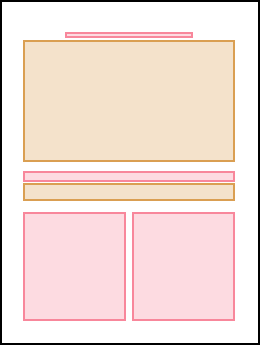} \\
        \includegraphics[width=0.2\textwidth]{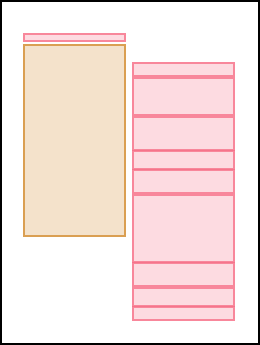} & \quad & \quad &\includegraphics[width=0.2\textwidth]{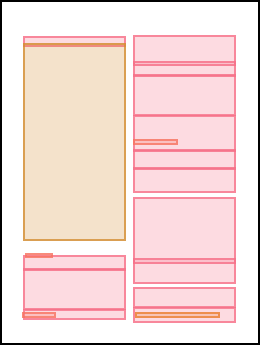} & 
        \includegraphics[width=0.2\textwidth]{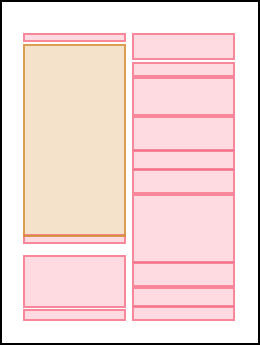} & \quad & \quad &\includegraphics[width=0.2\textwidth]{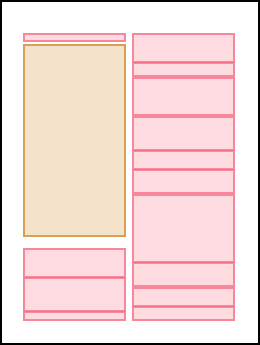} \\
        \includegraphics[width=0.2\textwidth]{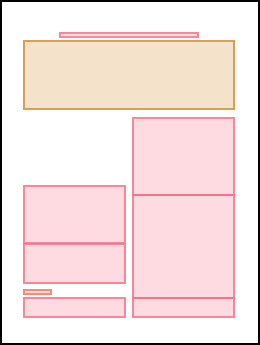} & \quad & \quad &\includegraphics[width=0.2\textwidth]{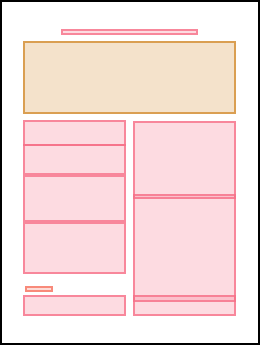} & 
        \includegraphics[width=0.2\textwidth]{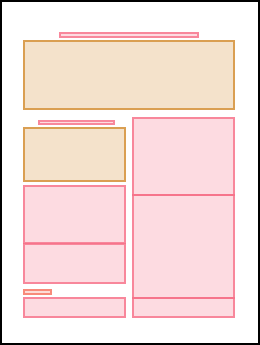} & \quad & \quad &\includegraphics[width=0.2\textwidth]{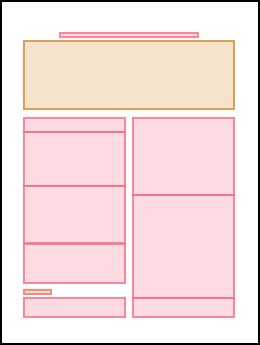} \\
        \includegraphics[width=0.2\textwidth]{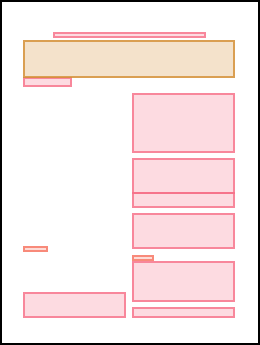} & \quad & \quad &\includegraphics[width=0.2\textwidth]{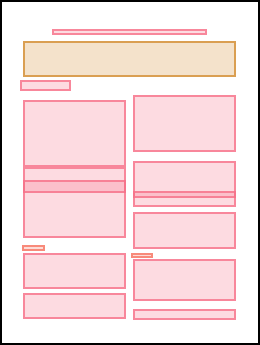} & 
        \includegraphics[width=0.2\textwidth]{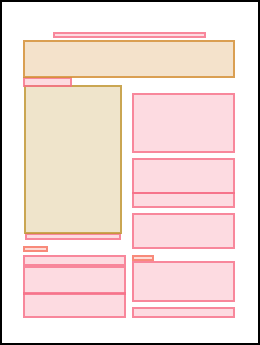} & \quad & \quad &\includegraphics[width=0.2\textwidth]{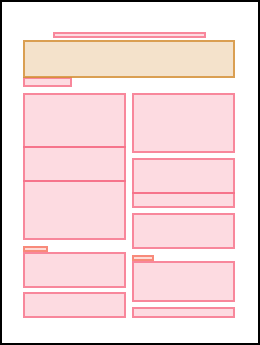} \\
        \includegraphics[width=0.2\textwidth]{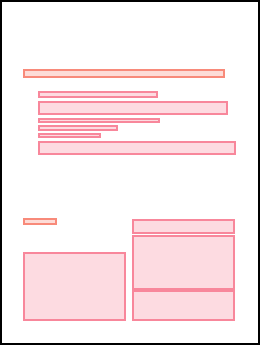} & \quad & \quad &\includegraphics[width=0.2\textwidth]{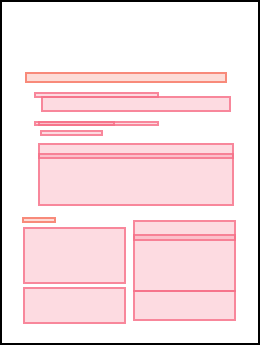} & 
        \includegraphics[width=0.2\textwidth]{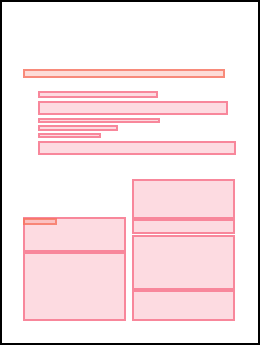} & \quad & \quad &\includegraphics[width=0.2\textwidth]{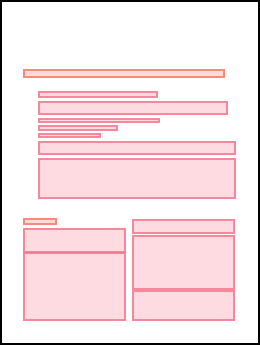} \\

    \end{tabular}
    \caption{Element Completion (0-20\%) on PubLayNet}
    
\end{table}

\begin{table}
    \centering
    \begin{tabular}{cc|cccc|cc}
        
        Input & \quad & \quad &  LayoutDM & LayoutFlow & \quad & \quad & Ground Truth \\
        \includegraphics[width=0.2\textwidth]{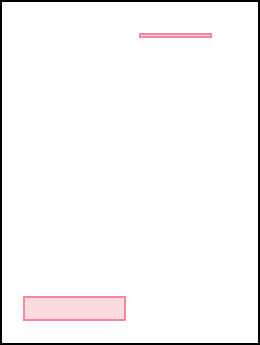} & \quad & \quad &\includegraphics[width=0.2\textwidth]{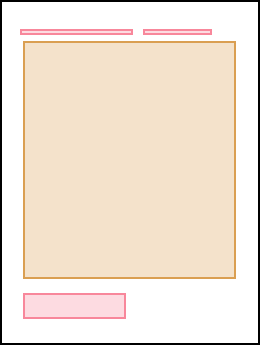} & 
        \includegraphics[width=0.2\textwidth]{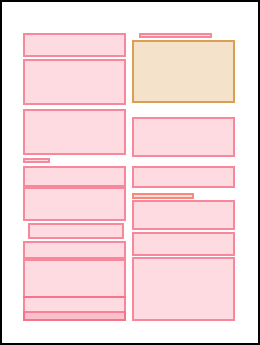} & \quad & \quad &\includegraphics[width=0.2\textwidth]{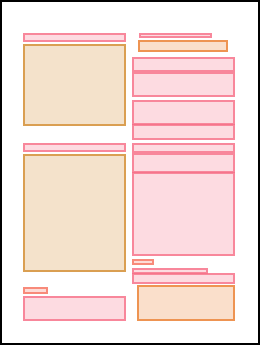} \\
        \includegraphics[width=0.2\textwidth]{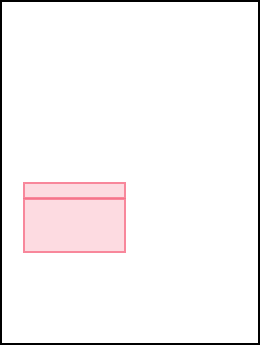} & \quad & \quad &\includegraphics[width=0.2\textwidth]{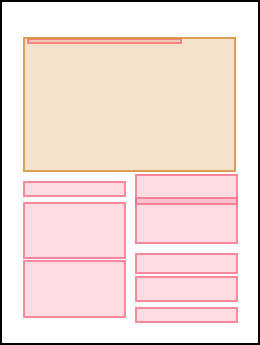} & 
        \includegraphics[width=0.2\textwidth]{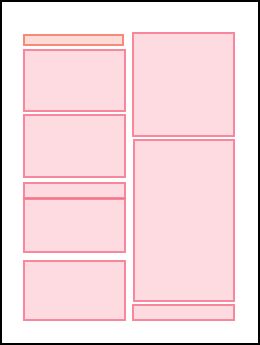} & \quad & \quad &\includegraphics[width=0.2\textwidth]{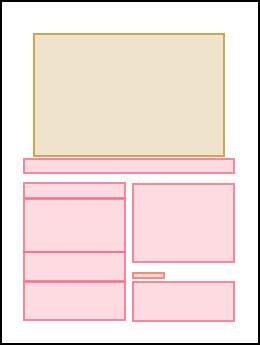} \\
        \includegraphics[width=0.2\textwidth]{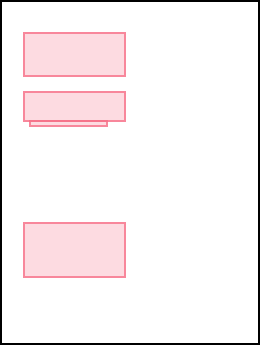} & \quad & \quad &\includegraphics[width=0.2\textwidth]{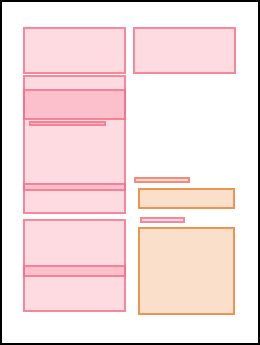} & 
        \includegraphics[width=0.2\textwidth]{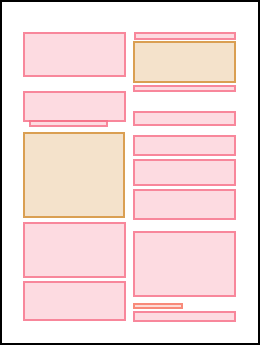} & \quad & \quad &\includegraphics[width=0.2\textwidth]{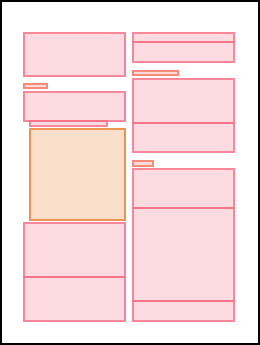} \\
        \includegraphics[width=0.2\textwidth]{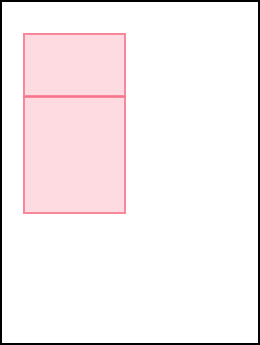} & \quad & \quad &\includegraphics[width=0.2\textwidth]{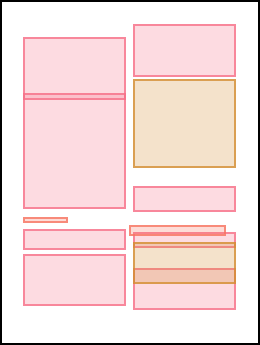} & 
        \includegraphics[width=0.2\textwidth]{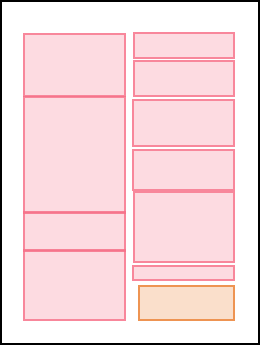} & \quad & \quad &\includegraphics[width=0.2\textwidth]{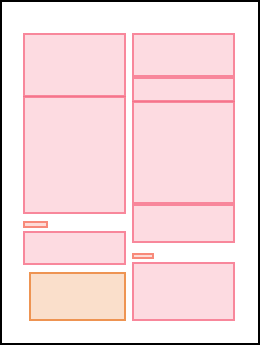} \\
        \includegraphics[width=0.2\textwidth]{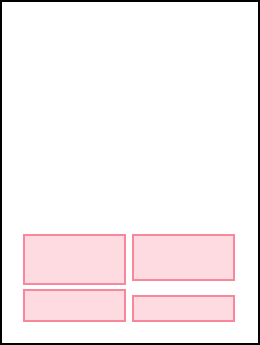} & \quad & \quad &\includegraphics[width=0.2\textwidth]{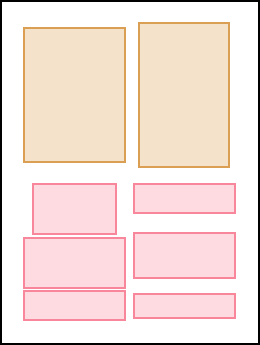} & 
        \includegraphics[width=0.2\textwidth]{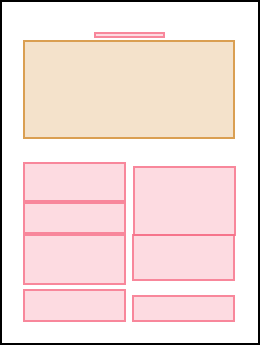} & \quad & \quad &\includegraphics[width=0.2\textwidth]{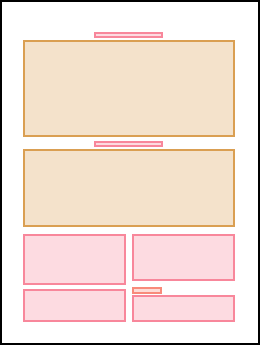} \\

    \end{tabular}
    \caption{Element Completion (80-100\%) on PubLayNet}
    
\end{table}

\begin{table}
    \centering
    \begin{tabular}{cc|ccc|cc}
        
        Input & \quad & \quad & LayoutFlow & \quad & \quad & Ground Truth \\
        \includegraphics[width=0.2\textwidth]{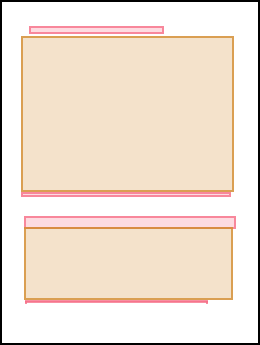} 
        & \quad & \quad &
        \includegraphics[width=0.2\textwidth]{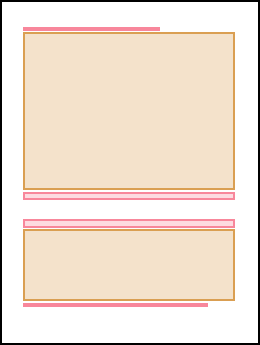} 
        & \quad & \quad &
        \includegraphics[width=0.2\textwidth]{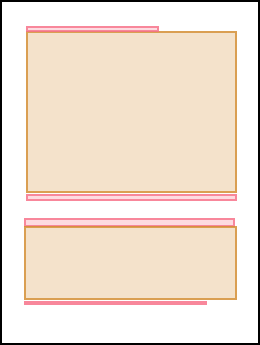} \\
        \includegraphics[width=0.2\textwidth]{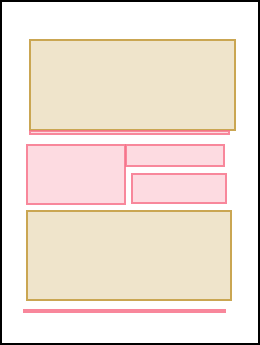} 
        & \quad & \quad &
        \includegraphics[width=0.2\textwidth]{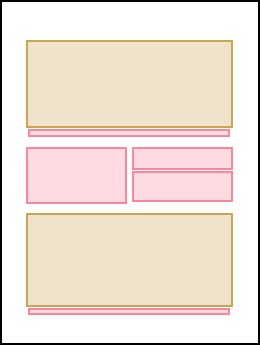} 
        & \quad & \quad &
        \includegraphics[width=0.2\textwidth]{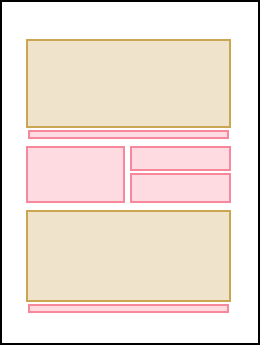} \\
        \includegraphics[width=0.2\textwidth]{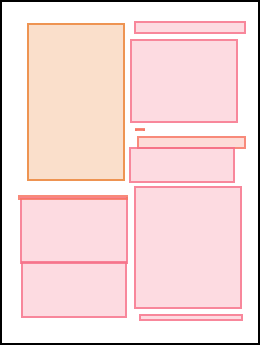} 
        & \quad & \quad &
        \includegraphics[width=0.2\textwidth]{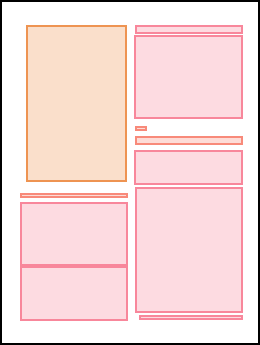} 
        & \quad & \quad &
        \includegraphics[width=0.2\textwidth]{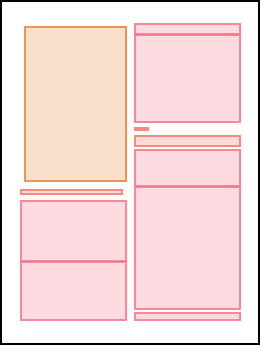} \\
        \includegraphics[width=0.2\textwidth]{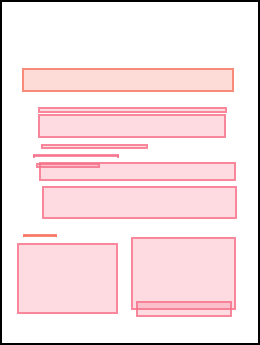} 
        & \quad & \quad &
        \includegraphics[width=0.2\textwidth]{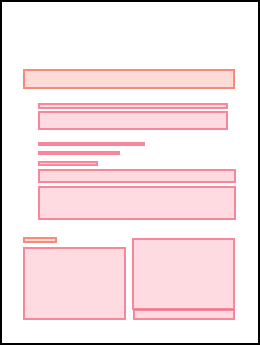} 
        & \quad & \quad &
        \includegraphics[width=0.2\textwidth]{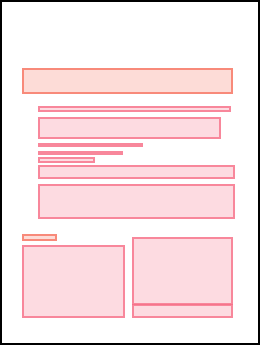} \\
        \includegraphics[width=0.2\textwidth]{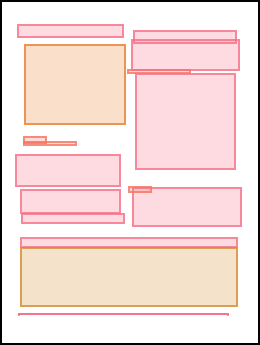} 
        & \quad & \quad &
        \includegraphics[width=0.2\textwidth]{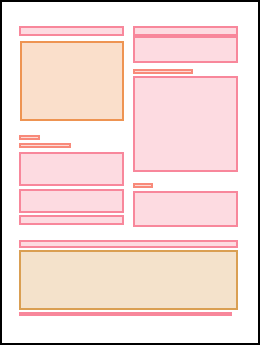} 
        & \quad & \quad &
        \includegraphics[width=0.2\textwidth]{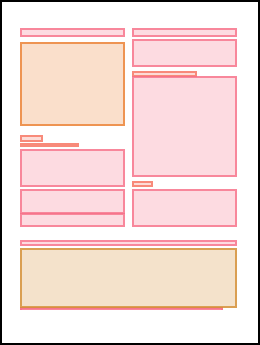} \\

    \end{tabular}
    \caption{Refinement on PubLayNet}
    
\end{table}
\end{center}

\begin{center}
    \begin{table}
    \centering
    \begin{tabular}{cc|cccc|cc}
        
        Input & \quad & \quad &  LayoutDM & LayoutFlow & \quad & \quad & Ground Truth \\
        \includegraphics[width=0.2\textwidth]{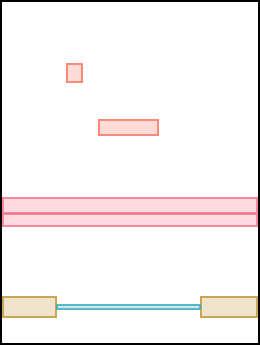} & \quad & \quad &\includegraphics[width=0.2\textwidth]{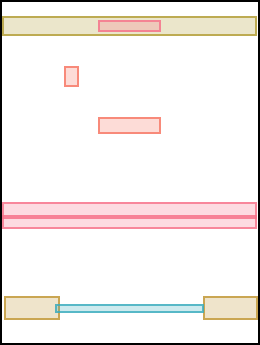} & 
        \includegraphics[width=0.2\textwidth]{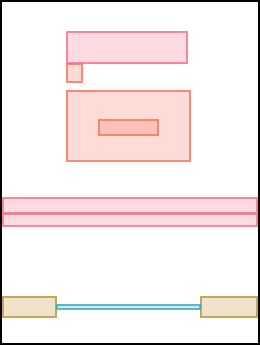} & \quad & \quad &\includegraphics[width=0.2\textwidth]{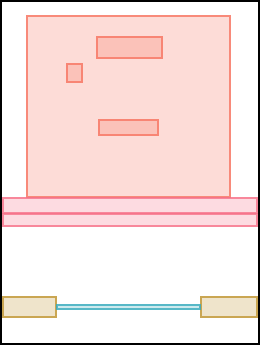} \\
        \includegraphics[width=0.2\textwidth]{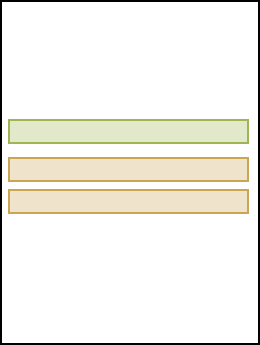} & \quad & \quad &\includegraphics[width=0.2\textwidth]{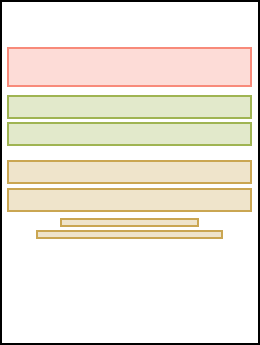} & 
        \includegraphics[width=0.2\textwidth]{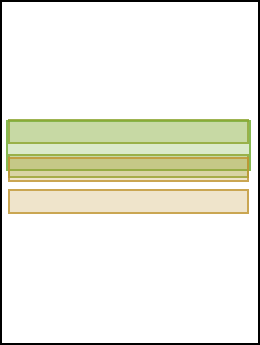} & \quad & \quad &\includegraphics[width=0.2\textwidth]{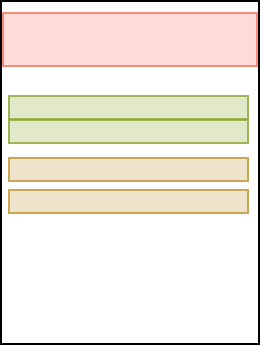} \\
        \includegraphics[width=0.2\textwidth]{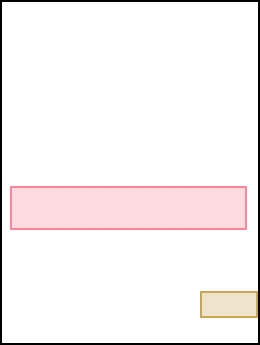} & \quad & \quad &\includegraphics[width=0.2\textwidth]{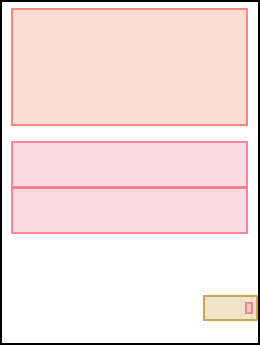} & 
        \includegraphics[width=0.2\textwidth]{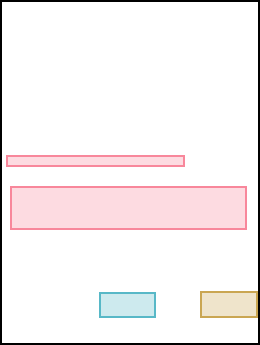} & \quad & \quad &\includegraphics[width=0.2\textwidth]{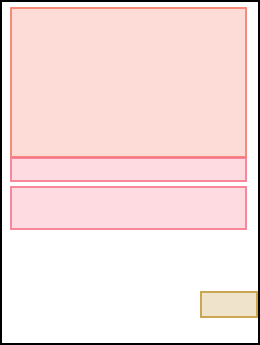} \\
        \includegraphics[width=0.2\textwidth]{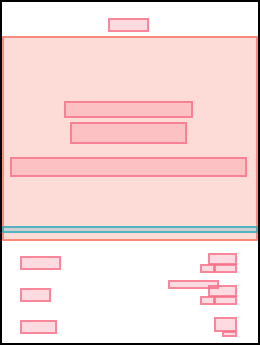} & \quad & \quad &\includegraphics[width=0.2\textwidth]{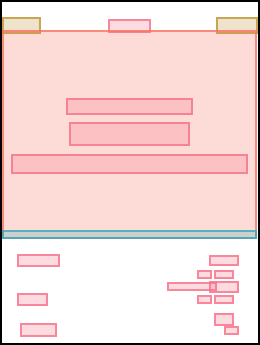} & 
        \includegraphics[width=0.2\textwidth]{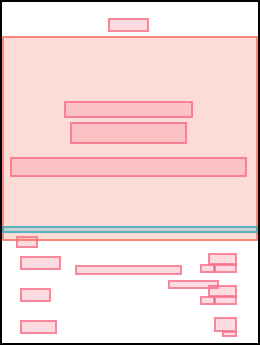} & \quad & \quad &\includegraphics[width=0.2\textwidth]{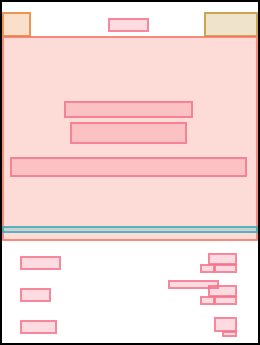} \\
        \includegraphics[width=0.2\textwidth]{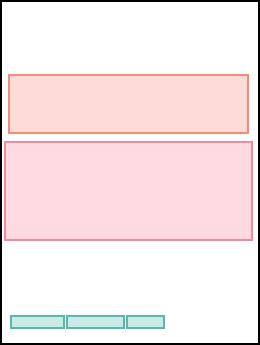} & \quad & \quad &\includegraphics[width=0.2\textwidth]{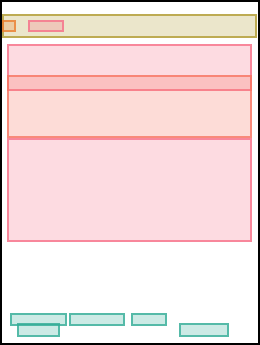} & 
        \includegraphics[width=0.2\textwidth]{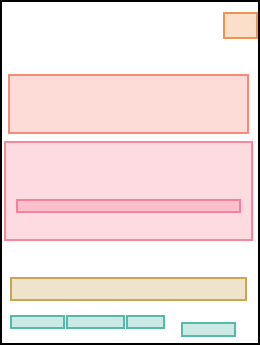} & \quad & \quad &\includegraphics[width=0.2\textwidth]{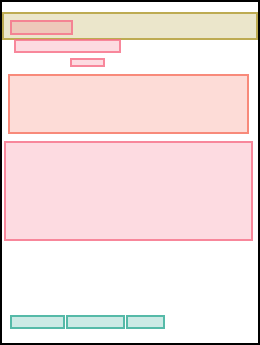} \\

    \end{tabular}
    \caption{Element Completion (0-20\%) on RICO}
    
\end{table}

\begin{table}
    \centering
    \begin{tabular}{cc|cccc|cc}
        
        Input & \quad & \quad &  LayoutDM & LayoutFlow & \quad & \quad & Ground Truth \\
        \includegraphics[width=0.2\textwidth]{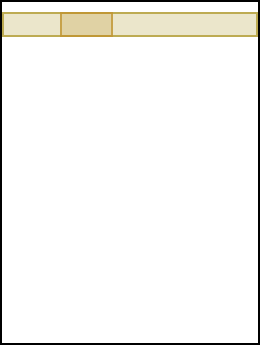} & \quad & \quad &\includegraphics[width=0.2\textwidth]{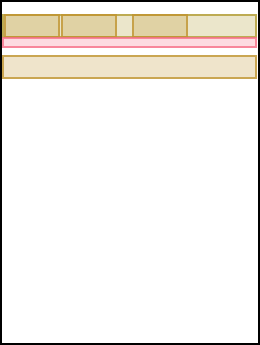} & 
        \includegraphics[width=0.2\textwidth]{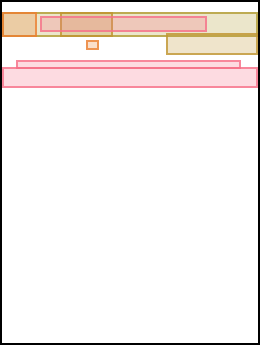} & \quad & \quad &\includegraphics[width=0.2\textwidth]{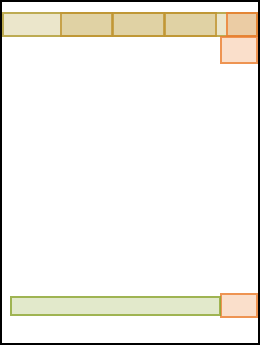} \\
        \includegraphics[width=0.2\textwidth]{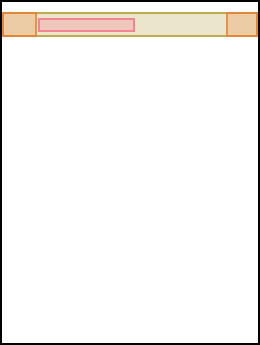} & \quad & \quad &\includegraphics[width=0.2\textwidth]{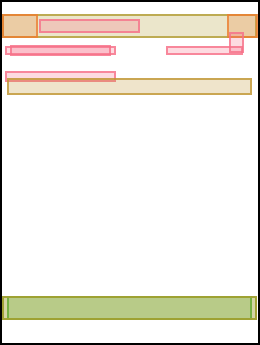} & 
        \includegraphics[width=0.2\textwidth]{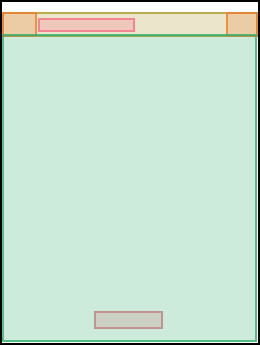} & \quad & \quad &\includegraphics[width=0.2\textwidth]{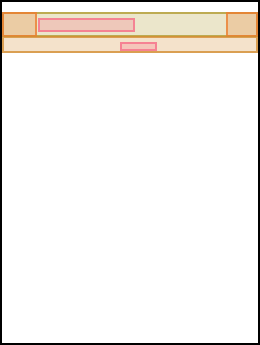} \\
        \includegraphics[width=0.2\textwidth]{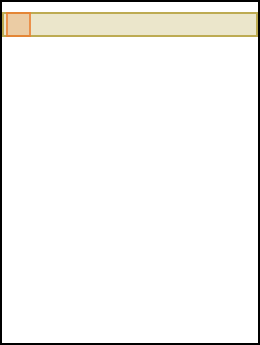} & \quad & \quad &\includegraphics[width=0.2\textwidth]{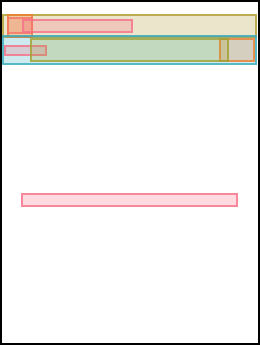} & 
        \includegraphics[width=0.2\textwidth]{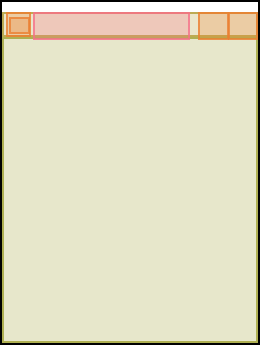} & \quad & \quad &\includegraphics[width=0.2\textwidth]{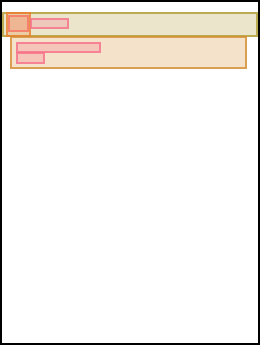} \\
        \includegraphics[width=0.2\textwidth]{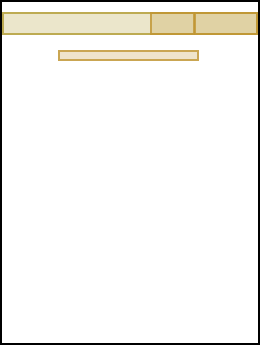} & \quad & \quad &\includegraphics[width=0.2\textwidth]{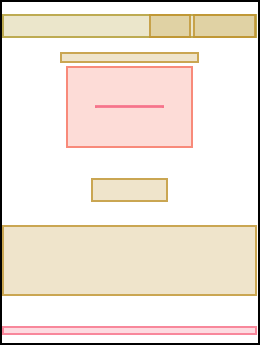} & 
        \includegraphics[width=0.2\textwidth]{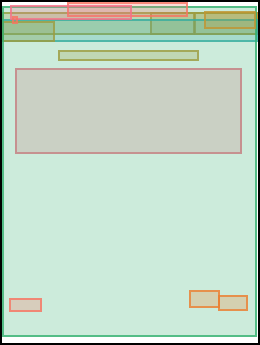} & \quad & \quad &\includegraphics[width=0.2\textwidth]{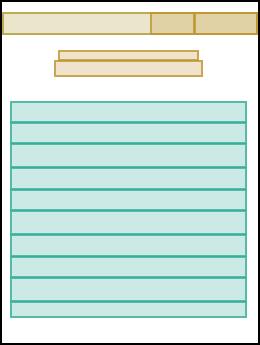} \\
        \includegraphics[width=0.2\textwidth]{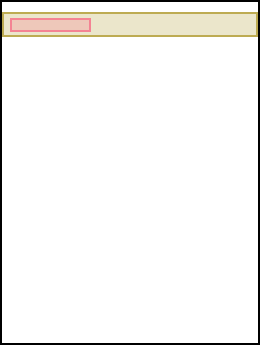} & \quad & \quad &\includegraphics[width=0.2\textwidth]{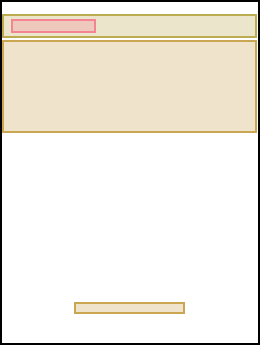} & 
        \includegraphics[width=0.2\textwidth]{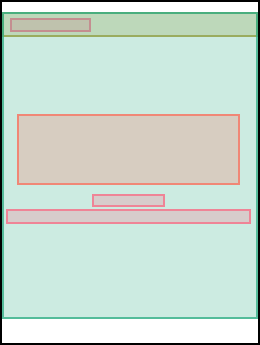} & \quad & \quad &\includegraphics[width=0.2\textwidth]{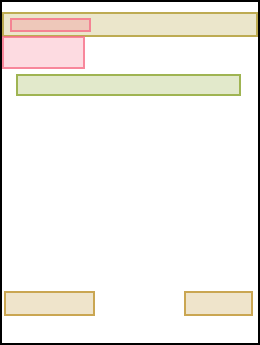} \\

    \end{tabular}
    \caption{Element Completion (80-100\%) on RICO}
    
\end{table}

\begin{table}
    \centering
    \begin{tabular}{cc|cccccc|cc}
        
        Input & \quad & \quad &  LayoutDM & LayoutDiffusion & LayoutFlow & \quad & \quad & Ground Truth \\
        \includegraphics[width=0.18\textwidth]{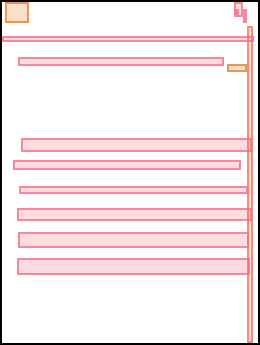} & 
        \quad & \quad &
        \includegraphics[width=0.18\textwidth]{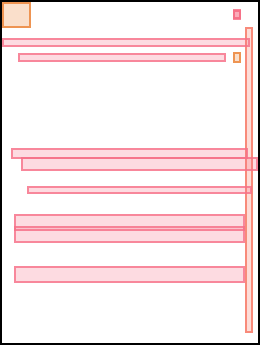} & 
        \includegraphics[width=0.18\textwidth]{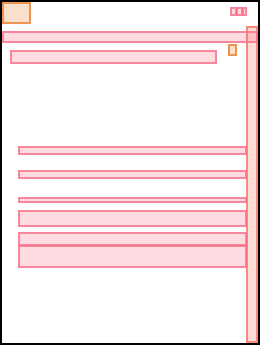} & 
        \includegraphics[width=0.18\textwidth]{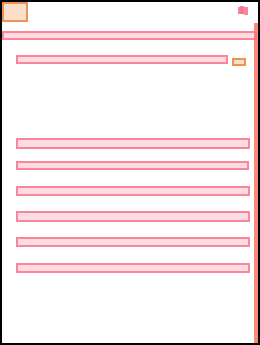} & 
        \quad & \quad &
        \includegraphics[width=0.18\textwidth]{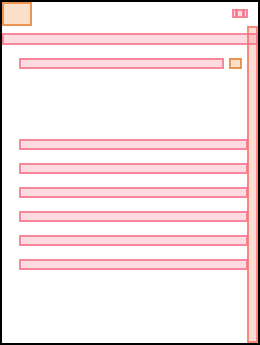} \\
        \includegraphics[width=0.18\textwidth]{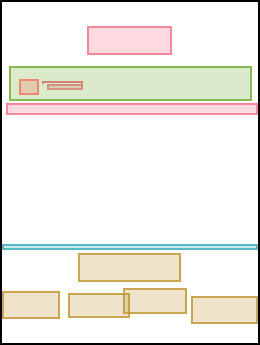} & 
        \quad & \quad &
        \includegraphics[width=0.18\textwidth]{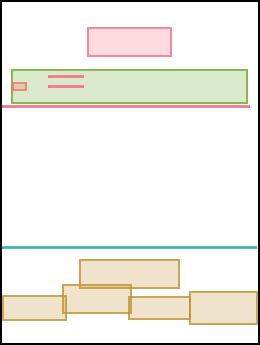} & 
        \includegraphics[width=0.18\textwidth]{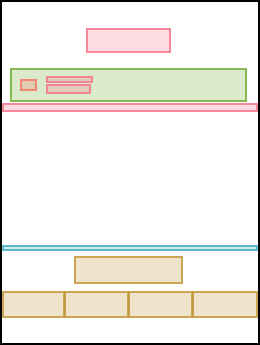} & 
        \includegraphics[width=0.18\textwidth]{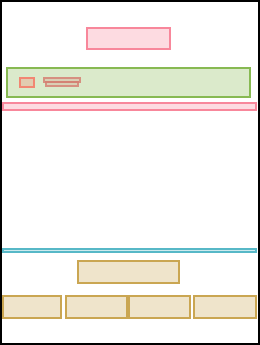} & 
        \quad & \quad &
        \includegraphics[width=0.18\textwidth]{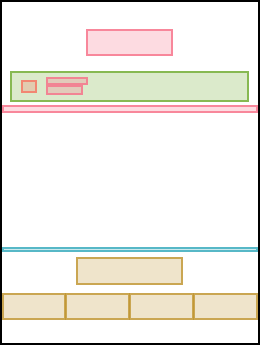} \\
        \includegraphics[width=0.18\textwidth]{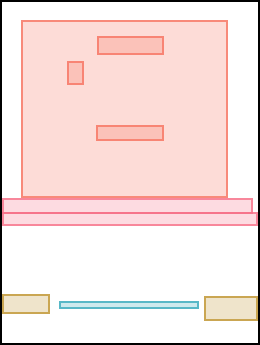} & 
        \quad & \quad &
        \includegraphics[width=0.18\textwidth]{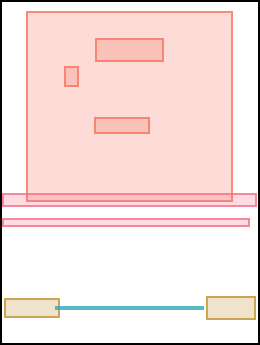} & 
        \includegraphics[width=0.18\textwidth]{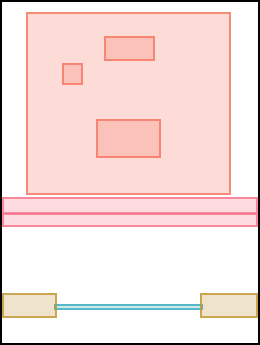} & 
        \includegraphics[width=0.18\textwidth]{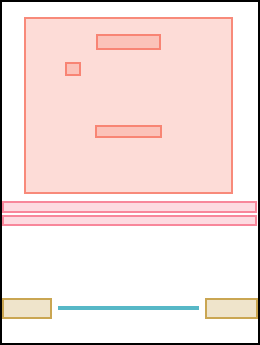} & 
        \quad & \quad &
        \includegraphics[width=0.18\textwidth]{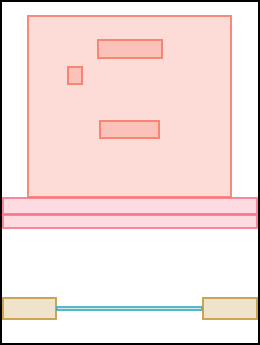} \\
        \includegraphics[width=0.18\textwidth]{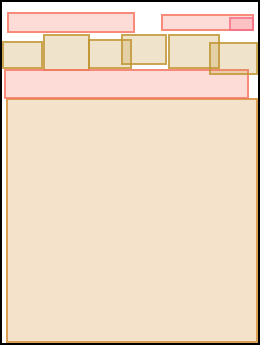} & 
        \quad & \quad &
        \includegraphics[width=0.18\textwidth]{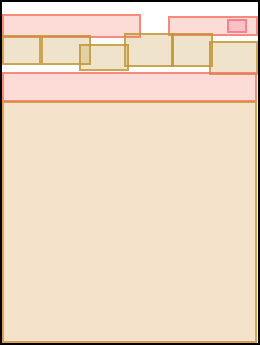} & 
        \includegraphics[width=0.18\textwidth]{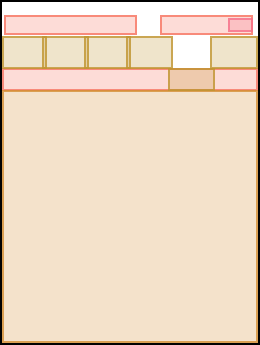} & 
        \includegraphics[width=0.18\textwidth]{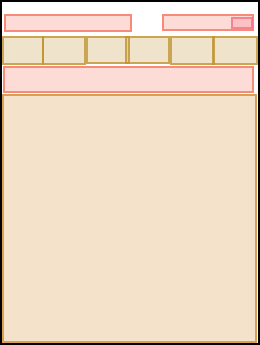} & 
        \quad & \quad &
        \includegraphics[width=0.18\textwidth]{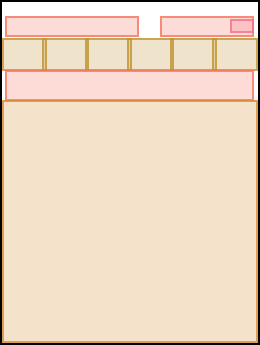} \\
        \includegraphics[width=0.18\textwidth]{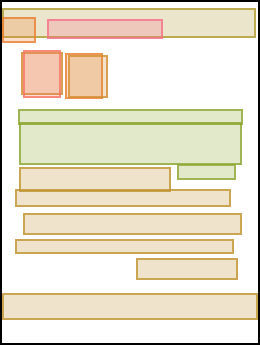} & 
        \quad & \quad &
        \includegraphics[width=0.18\textwidth]{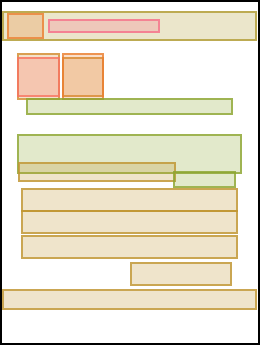} & 
        \includegraphics[width=0.18\textwidth]{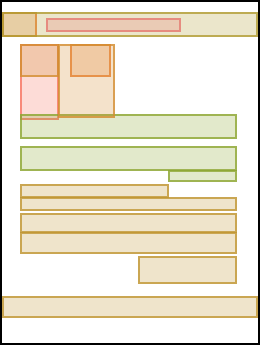} & 
        \includegraphics[width=0.18\textwidth]{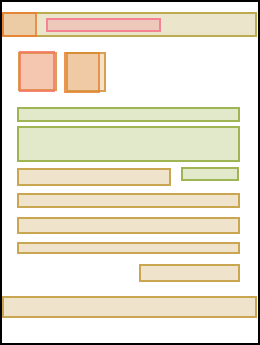} & 
        \quad & \quad &
        \includegraphics[width=0.18\textwidth]{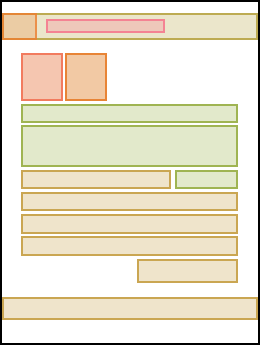} \\

    \end{tabular}
    \caption{Refinement on RICO}
    
\end{table}
\end{center}
\end{document}